\documentclass[12pt]{article}
\usepackage[utf8]{inputenc}

\usepackage[top=1in, bottom=1in, left=1in, right=1in]{geometry}
\usepackage{graphicx}
\usepackage{subfig}
\usepackage{amsthm}
\usepackage[comma,numbers]{natbib}
\usepackage{ulem}
\usepackage{color}
\usepackage{lineno}
\usepackage{hyperref}
\usepackage{amsmath}
\usepackage{amssymb}
\usepackage{bm}
\usepackage{indentfirst}
\usepackage{mathrsfs}
\usepackage{algorithm}
\usepackage{algpseudocode}
\usepackage{threeparttable}
\usepackage{booktabs}
\usepackage{multirow}

\usepackage{hyperref}
\hypersetup{
hypertex=true,
colorlinks=true,
linkcolor=blue,
anchorcolor=blue,
citecolor=blue
}
\newtheorem{theorem}{Theorem}

\newcommand\keywords[1]{\textbf{Keywords}: #1}

\title{Identifying parameter couplings and uncertainties of mixed-noise stochastic systems via full-covariance Gaussian mixture network}

\author{\small{Xiaolong Wang$^{1,2,3}$, Xiangwen Hao$^{1}$, Jing Feng$^4$, Yuanyuan Liu$^4$, Yong Xu$^{2,3}$\footnote{Corresponding author. E-mail addresses:  {\it hsux3@nwpu.edu.cn} (Y. Xu)}} \\
\small{$^1$ School of Mathematics and Statistics,}\\ \small{Shaanxi Normal University, Xi'an, 710119, China}\\
\small{$^2$ School of Mathematics and Statistics,}\\ \small{Northwestern Polytechnical University, Xi'an, 710129, China}\\
\small{$^3$ MOE Key Laboratory for Complexity Science in Aerospace,}\\ \small{Northwestern Polytechnical University, Xi'an, 710072, China}\\
\small{$^4$ School of Science, Xi'an University of Posts and Telecommunications,}\\ \small{Xi'an, 710121, China}
}
\date{}

\begin{document}

\maketitle


\begin{abstract}
Parameter identification of stochastic dynamical systems driven by mixed noises is challenging due to intractable likelihood functions. We propose PENN-GMD, a parameter estimation neural network that maps partially observed trajectories to a Gaussian mixture distribution (GMD) over the system parameters. Unlike conventional uncertainty estimates, the GMD employs full covariance matrices to explicitly reveal parameter couplings and multi-modal likelihood structures. The network is trained by minimizing the negative log-likelihood via a surjective parameterization that hard-encodes all GMD constraints, thereby approximating the true likelihood. We validate the method on five numerical examples with increasing complexity, including systems driven by fractional Gaussian and L{\'e}vy noises, oscillators with colored noise, coupled neurons under different observability, and an aeroelastic airfoil with unidentifiable stochastic disturbances. Results demonstrate that PENN-GMD accurately recovers likelihood distributions, captures parameter couplings, and naturally diagnoses non-identifiability through variance broadening or mode splitting. These capabilities establish PENN-GMD as a practical tool for uncertainty-aware parameter identification in complex stochastic systems where conventional likelihood-based methods are infeasible.
\bigskip

\noindent\keywords{nonlinear stochastic system, parameter identification, uncertainty quantification, deep learning, Gaussian mixture distribution.}
\end{abstract}

\section{Introduction}
\label{sec:intro}
Stochastic dynamical systems are widely employed to model real-world physical systems subject to noisy disturbances~\cite{McDonnell2011benefits,Palmer2019Stochastic}. Accurate parameter identification of the deterministic system and the driving noises from measurement data is essential for understanding and predicting the stochastic dynamics. In many practical scenarios, the stochastic disturbance is not a single noise source but a superposition of multiple independent drivers~\cite{Zan2020Stochastic,Mi2023Stochastic,Qu2025Existence}. The coupling between nonlinearities and mixed noises induces complex structures in the parameter space, including strong couplings among parameters, pronounced uncertainties, and even complete non-identifiability. These challenges render point estimates insufficient. When the data do not uniquely determine the parameters, an honest quantification of estimation uncertainty becomes essential. The goal of this work is to recover the likelihood distribution from a single observed trajectory, which encodes all information about which parameter combinations are consistent with the observed data. By moving beyond point estimates, we can capture the couplings among parameters and the uncertainties inherent in the estimation process, revealing the complex coupling structures that arise in mixed-noise driven stochastic systems.

Traditional likelihood-based methods, such as maximum likelihood estimation (MLE)~\cite{Parameter2000,Bishwal2007ParameterEI}, require an explicit and analytically tractable likelihood function of the noise. This requirement, however, becomes problematic for many practically relevant noise types. Fractional Gaussian noise (fGn)~\cite{Mandelbrot1968Fractional}, for instance, captures long-range dependence with a covariance function that decays polynomially~\cite{Weiss2013Single,Benelli2021From}, governed by the Hurst exponent. L{\'e}vy noise characterizes heavy-tailed fluctuations with intermittent large jumps controlled by the stability index~\cite{Sato1999LvyPA}, offering a natural description of extreme events~\cite{Feng2018Phase,Zan2021First,Zhang2021Rate}. Despite the theoretical appeal and modeling flexibility of these non-Gaussian or non-white noise models, they pose significant computational challenges for likelihood-based inference. For fGn, the covariance matrix is dense and its inverse must be recomputed at each likelihood evaluation~\cite{Feng2023Deep}, leading to substantial computational cost. For L{\'e}vy noise, no closed-form probability density function (PDF) exists for general stability indices, and the likelihood must be evaluated through expensive numerical integration~\cite{Wang2022Neural}. In such situations, least squares estimator~\cite{Hu2009Least}, moment estimator~\cite{Cheng2020Generalized}, and the MLE~\cite{Parameter2017,Wei2022ParameterEF,Masuda2019Non} are often restricted to identify linear systems, low-dimensional systems or systems with specific structures. When the system is multi-dimensional, controlled by multiple parameters, driven by multiple noise sources and only partially observed, the construction of the likelihood function becomes even more challenging and often infeasible. Consequently, there is a clear need for methods that bypass explicit likelihood specification while still providing a principled quantification of parameter uncertainty.

The recent advent of deep learning has opened new avenues for solving inverse problems, as deep neural networks can extract complex, high-level features from raw data~\cite{Nah2017Deep,Zhu2018Image}. In the study of anomalous diffusion, neural networks have been shown to outperform traditional mean squared displacement analysis in estimating the Hurst exponent and classifying the underlying diffusion model from single-particle trajectories~\cite{MuozGil2021ObjectiveCO}. These studies have inspired a broader simulation-based learning paradigm for parameter estimation in stochastic systems. The key idea is to reformulate the inverse problem, i.e., inferring parameters from an observed trajectory, as a supervised learning task from simulated forward data. First, a large corpus of trajectories is efficiently generated from known stochastic differential equations (SDEs) by simulation, covering diverse parameter choices. Second, a neural network is trained to learn the inverse mapping from data to parameters. The intricate mapping between raw trajectories and the full parameters of the stochastic systems is learned in the training process without explicit likelihood formulations.

The original parameter estimation neural network (PENN) framework was proposed for joint estimation of system and noise parameters in SDEs driven by L{\'e}vy noise from a single discretely sampled trajectory~\cite{Wang2022Neural}. Subsequent extensions have generalized this idea to identify SDEs driven by fGn~\cite{Feng2023Deep}, L{\'e}vy colored noise~\cite{Wang2025Noise}, Student L{\'e}vy processes~\cite{Li2024Parameter}, and time-varying SDEs with partially observed states~\cite{Feng2025Fusing}. The PENN architecture combines a long short-term memory (LSTM) encoder for trajectory feature extraction and a fully-connected neural network (FCNN) decoder for mapping these features to the system parameters. Convolutional layers, attention mechanisms, and transformers~\cite{Hou2025Deep,Hou2026PVNN} have also been integrated into the framework to improve efficiency. The original PENNs are point estimators trained with weighted mean absolute error (MAE)~\cite{Wang2022Neural} or mean squared error~\cite{Feng2023Deep} between the true parameter vectors and the estimates. Improved PENNs~\cite{Wang2025Noise,Feng2025Fusing} model the system parameters using a multivariate Gaussian distribution with a diagonal covariance matrix and minimize the negative log-likelihood. This enhancement enables the PENNs not only to obtain parameter estimates (the means) but also to quantify the uncertainty of these estimates via the predicted standard deviation (SD). While a single diagonal Gaussian remains a popular uncertainty quantification strategy~\cite{Lu2025Structural,Bae2025Inferring}, it cannot capture parameter couplings and is incapable of representing multi-modal likelihood structures. 

The PENN framework is closely related to the amortized inference paradigm, where a computationally expensive training phase is performed once using simulated data, after which inference for any new observation requires only forward passes through the trained network. BayesFlow~\cite{Radev2022BayesFlow} and MINIMALIST~\cite{Isacchini2022Mutual} are two representative amortized methods that learn implicit posterior representations, respectively via a normalizing flow and an energy function. In both cases, obtaining posterior samples requires repeated forward evaluations, either through latent Gaussian sampling or MCMC, which limits the fully amortized nature of the inference. More importantly, their implicit representations do not directly provide uncertainty quantification, which must be inferred from post-hoc sample analysis. These limitations motivate the approach proposed below.

In this work, we propose PENN-GMD, an end-to-end neural network estimator that maps single trajectories into likelihood distributions of the system parameters, which are expressed by Gaussian mixture distributions (GMD). This yields a closed-form likelihood with full covariance matrices, providing immediate access to parameter couplings and multi-modal structures that arise from non-identifiability or limited data. The novelty is threefold.

1. We extend the PENN framework by building a map from a trajectory to a GMD with full covariance matrices, where all distributional constraints are hard-encoded via Cholesky decomposition and softmax/softplus transformations. This design leads to a single-term learning objective without constraint-related hyperparameters to tune, resulting in an efficient and stable training process.

2. We present an analysis of why the method outputs likelihood distributions and numerically verify the close alignment between PENN-GMD and the theoretical likelihood expression on a Gaussian-driven Ornstein-Uhlenbeck process.

3. Through four mixed-noise examples spanning linear to strongly nonlinear systems, Gaussian to heavy-tailed non-Gaussian noises, and full to partial observability, we demonstrate that PENN-GMD not only recovers accurate parameter estimates but also reveals physically interpretable parameter couplings, and faithfully diagnoses non-identifiability through the multi-modal structure of the GMD. These capabilities are inaccessible to conventional point estimators or diagonal uncertainty models.

The remainder of this paper is organized as follows. Sec.~\ref{sec:definition} formulates the problem. Sec.~\ref{sec:penn} details the proposed PENN-GMD framework. Sec.~\ref{sec:numer_ex} presents extensive numerical experiments. Sec.~\ref{sec:discussion} discusses the limitations and potential extensions of the approach via an ablation study. Finally, Sec.~\ref{sec:conclusion} concludes the paper.

\section{Problem definition}
\label{sec:definition}
We consider a $D$-dimensional nonlinear stochastic system described by known SDEs with state $\mathbf{x}(t)\in\mathbb{R}^D$ and $M$ system parameters $\boldsymbol{\Theta}^*=[\theta_1^*,\dots,\theta_M^*]^\top\in \mathcal{P} \subset \mathbb{R}^M$. The parameter domain $\mathcal{P}$ is a hyperrectangle
\begin{equation}
\mathcal{P} = \left\{ \boldsymbol{\Theta} = [\theta_1,\dots,\theta_M]^\top \;\middle|\; \theta_i \in [\theta_i^{\min},\, \theta_i^{\max}],\ i=1,\dots,M \right\},
\end{equation}
which covers all possible parameter configurations of interest.

Using numerical integration such as the Euler-Maruyama method, we can simulate $\mathbf{x}(t)$ from an initial state $\mathbf{x}(0)$ for a given parameter vector $\boldsymbol{\Theta}^*\in\mathcal{P}$ with sampling time $\Delta t$. The simulation produces a trajectory $\mathbf{X} = \{\mathbf{x}_i\}_{i=1}^L \triangleq \mathbf{x}_{1:L}$ of length $L$, where $\mathbf{x}_i = \mathbf{x}(t_0 + (i-1)\Delta t)$ and $t_0$ is chosen sufficiently large to remove transient effects. In many practical applications, however, the state $\mathbf{x}(t)$ cannot be fully observed. Instead, only partial states or their combinations are available. We define a measurement function $\mathbf{z} = Q(\mathbf{x})$ and call $\{\mathbf{z}_i\}_{i=1}^L \triangleq \mathbf{z}_{1:L}$ an observed trajectory, with $\mathbf{z}_i = Q(\mathbf{x}_i)$.

Since the system equations are known, we can generate a training dataset $\mathcal{D} = \{(\mathbf{Z}_j, \boldsymbol{\Theta}_j^*)\}_{j=1}^{N_{\text{train}}}$ consisting of $N_{\text{train}}$ trajectories. The $j$-th trajectory is generated by first uniformly sampling $\boldsymbol{\Theta}_j^* \in \mathcal{P}$, then simulating and recording the observed trajectory $\mathbf{Z}_j = \mathbf{z}_{j,1:L}$. The dataset $\mathcal{D}$ thus covers the entire parameter domain $\mathcal{P}$.

The goal of this work is to build a PENN that, given an arbitrary test trajectory $\mathbf{Z}=\mathbf{z}_{1:n}$ of length $n$, infers the likelihood distribution $p(\mathbf{Z} \mid \boldsymbol{\Theta})$ of the underlying system parameters $\boldsymbol{\Theta}$. In the test stage, the PENN outputs a parameter vector $\boldsymbol{\Phi}$
\begin{equation}
\boldsymbol{\Phi} = \text{PENN}(\mathbf{Z}),
\end{equation}
which defines a parameterized distribution $q(\boldsymbol{\Theta};\boldsymbol{\Phi})$ that approximates the likelihood function $p(\mathbf{Z} \mid \boldsymbol{\Theta})$. The covariance and multi-modality of $q$ reveal the couplings and uncertainties among the system parameters $\boldsymbol{\Theta}$.

The uniform sampling of $\mathcal{P}$ reflects the belief that all parameter choices are equally important. From a Bayesian perspective, it corresponds to a uniform prior $p(\boldsymbol{\Theta}) \propto 1$ within the parameter domain. This assumption implies that the posterior $p(\boldsymbol{\Theta} \mid \mathbf{Z})$ is proportional to the likelihood $p(\mathbf{Z} \mid \boldsymbol{\Theta})$:
\begin{equation}
p(\boldsymbol{\Theta} \mid \mathbf{Z})
= \frac{p(\mathbf{Z} \mid \boldsymbol{\Theta}) \, p(\boldsymbol{\Theta})}{p(\mathbf{Z})}
\propto p(\mathbf{Z} \mid \boldsymbol{\Theta}).
\end{equation}

This interpretation motivates us to train the PENN by minimizing the empirical negative log-likelihood (NLL) of the trajectories and corresponding true parameters in the training dataset $\mathcal{D}$~\cite{PRML}:
\begin{equation}\label{eq:emp_and_kld}
\begin{split}
\mathcal{L}_\text{emp} &= -\frac{1}{N_{\text{train}}}\sum_{j=1}^{N_{\text{train}}} \ln q(\boldsymbol{\Theta}_j^* \mid \mathbf{Z}_j)\\
&\approx -\mathbb{E}_{p(\boldsymbol{\Theta},\mathbf{Z})}[\ln q(\boldsymbol{\Theta} \mid \mathbf{Z})]
= \int p(\boldsymbol{\Theta},\mathbf{Z}) \ln q(\boldsymbol{\Theta}\mid\mathbf{Z})\mathrm{d}\boldsymbol{\Theta}\mathrm{d}\mathbf{Z}\\
& = \int p(\boldsymbol{\Theta},\mathbf{Z}) \ln \frac{p(\boldsymbol{\Theta}\mid\mathbf{Z})}{q(\boldsymbol{\Theta}\mid\mathbf{Z})}\mathrm{d}\boldsymbol{\Theta}\mathrm{d}\mathbf{Z} - \int p(\boldsymbol{\Theta},\mathbf{Z}) \ln p(\boldsymbol{\Theta}\mid\mathbf{Z})\mathrm{d}\boldsymbol{\Theta}\mathrm{d}\mathbf{Z}\\
& = \int p(\mathbf{Z}) \left[ \int p(\boldsymbol{\Theta}\mid\mathbf{Z}) \ln \frac{p(\boldsymbol{\Theta}\mid\mathbf{Z})}{q(\boldsymbol{\Theta}\mid\mathbf{Z})} \mathrm{d}\boldsymbol{\Theta} \right] \mathrm{d}\mathbf{Z} + \text{const} \\
&= \mathbb{E}_{p(\mathbf{Z})}[\mathrm{KL}(p(\boldsymbol{\Theta}\mid\mathbf{Z}) \| q(\boldsymbol{\Theta}\mid\mathbf{Z}))] + \text{const}.
\end{split}
\end{equation}
where the constant is independent of the learning goal $q(\boldsymbol{\Theta};\boldsymbol{\Phi})$. The empirical NLL $\mathcal{L}_\text{emp}$ in the first line is a Monte Carlo approximation to the population-level NLL $-\mathbb{E}_{p(\boldsymbol{\Theta},\mathbf{Z})}[\ln q(\boldsymbol{\Theta} \mid \mathbf{Z})]$ in the second line, where the expectation is taken over the joint distribution of parameters and trajectories. By the law of large numbers, the approximation error vanishes as $N_{\text{train}} \to \infty$. The equalities in the last row further indicate that minimizing the population-level NLL is equivalent to minimizing the Kullback-Leibler divergence between $p(\boldsymbol{\Theta}\mid\mathbf{Z})$ and $q(\boldsymbol{\Theta}\mid\mathbf{Z})$, averaged over all possible trajectories. Thus, minimizing the NLL drives $q$ toward the true posterior $p(\boldsymbol{\Theta} \mid \mathbf{Z})$ for every trajectory, which, with the uniform prior, is proportional to the likelihood $p(\mathbf{Z} \mid \boldsymbol{\Theta})$. Consequently, the mixture distribution output by the PENN serves as a closed-form parametric approximation to the likelihood distribution, thereby obviating explicit likelihood evaluation at test time.

This interpretation offers a significant practical advantage. In traditional parameter estimation, constructing the likelihood function requires a fully specified probabilistic model of the system, including the noise distributions and their dependencies, which is often intractable for systems driven by multi-source non-Gaussian disturbances, non-white noises, or high-dimensional dynamics with only partial state observations. The proposed PENN bypasses this requirement entirely. It learns the likelihood directly from simulated trajectory-parameter pairs, without the need for explicit probabilistic modeling of the noise sources or the system dynamics. This greatly simplifies the uncertainty quantification workflow for complex stochastic systems, making it applicable to a wide range of problems where conventional likelihood-based methods are infeasible.

\section{Parameter estimation neural network with Gaussian mixture distribution}
\label{sec:penn}
\begin{figure}[!htb]
\center{\includegraphics[width=1\textwidth]
{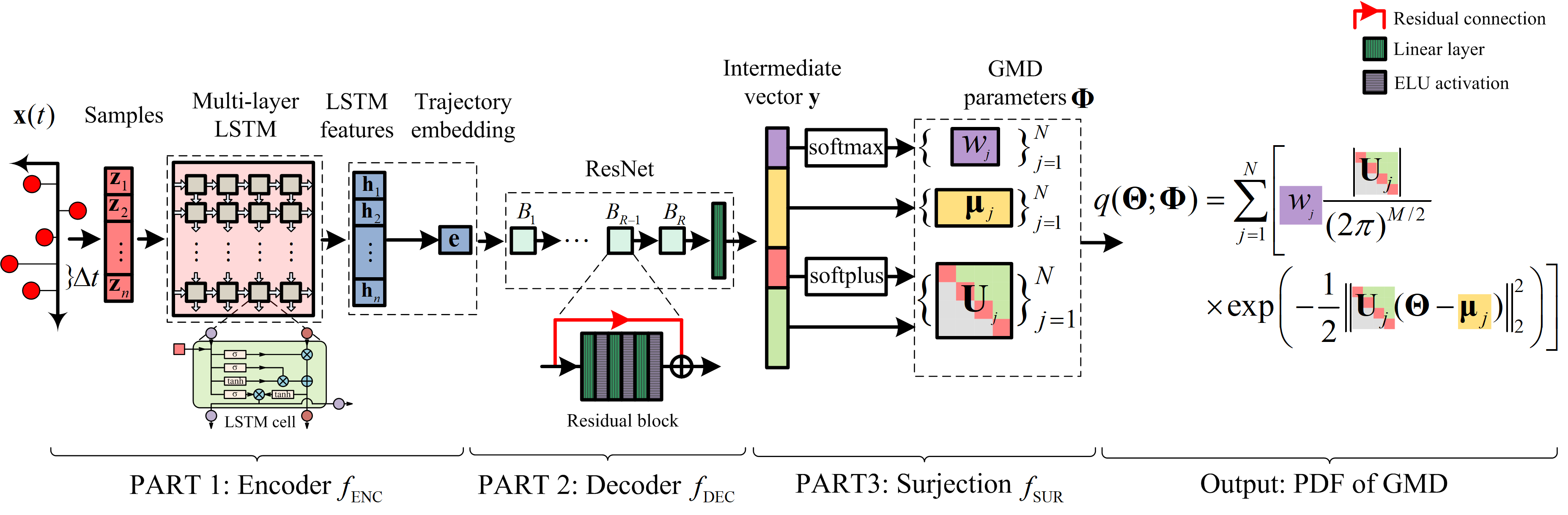}}
\caption{\label{fig:network_architecture}The architecture of the PENN-GMD framework.}
\end{figure}

The proposed PENN-GMD adopts an encoder-decoder architecture for parameter estimation. It consists of three modular components that map a sample trajectory to a GMD, as shown in Fig.~\ref{fig:network_architecture}. First, the encoder $f_{\text{ENC}}$ maps the observed trajectory $\mathbf{z}_{1:n}$ to a fixed-length embedding vector $\mathbf{e}\in\mathbb{R}^\text{EMB}$. Second, the decoder $f_{\text{DEC}}$ refines this embedding into an unconstrained vector $\mathbf{y}$. Finally, a surjective mapping $f_{\text{SUR}}$ recovers the feasible GMD parameters $\boldsymbol{\Phi}$ from $\mathbf{y}$, resulting in the parameter distribution $q(\boldsymbol{\Theta};\boldsymbol{\Phi})$ of the trajectory.

We first introduce the GMD and the surjection $f_{\text{SUR}}$ onto the parameter domain of the GMD in Sec.~\ref{sec:gmd}. Then the architectures of the encoder and decoder are detailed in Sec.~\ref{sec:nn}. The loss function is defined in Sec.~\ref{sec:loss} and the training algorithm is summarized in Sec.~\ref{sec:training}.

\subsection{Gaussian mixture distribution}
\label{sec:gmd}
In this work, we adopt the GMD $q(\boldsymbol{\Theta};\boldsymbol{\Phi})$ to represent the $M$-D likelihood distribution $p(\mathbf{Z}|\boldsymbol{\Theta})$ for several reasons. First, the GMD is capable of modeling multi-dimensional, multi-modal distributions, where parameter couplings and uncertainties are explicitly encoded in the covariance matrices of the mixture components. Second, the GMD is a universal approximator for smooth densities~\cite{Goodfellow2016Deep,Lindberg2025Estimating}, while its identifiability property~\cite{Delon2020Wasserstein} ensures that up to the trivial label-switching ambiguity, two distinct sets of GMD parameters cannot represent the same distribution, thereby avoiding solution ambiguity. Third, the GMD admits closed-form expressions for its mean and covariance matrix, and its low-dimensional marginal distributions remain GMDs, facilitating both interpretation and visualization, as will be extensively shown in Sec.~\ref{sec:numer_ex}. Finally, this section shows that the parameters of a GMD can be vectorized, making it a convenient output format for neural networks.

A key design choice in the GMD formulation is the use of full covariance matrices~\cite{He2020Deep}, as opposed to the more common diagonal covariance matrices~\cite{Munoz2025Quantitative,Yu2024Multi,Xie2023Incorporating,Wang2021Unsupervised}. When the number of components is fixed, diagonal-covariance GMDs are inefficient at representing strongly correlated likelihood functions, as capturing a tilted ridge requires many axis-aligned components. This limitation is especially severe when identifying systems with many parameters, where the parameter domain is high-dimensional. Full covariance matrices overcome this by aligning with the couplings direction using fewer components, offering a more accurate representation when the component budget is fixed. For these reasons, we adopt full covariance matrices in this work.

Formally, a $M$-D GMD with $N$ components is defined by
\begin{equation}\label{eq:gmd_density}
q(\boldsymbol{\Theta};\boldsymbol{\Phi})=\sum_{j=1}^{N} w_j \mathcal{N}\left(\boldsymbol{\Theta};\boldsymbol{\mu}_j,\boldsymbol{\Sigma}_j\right),
\end{equation}
where the GMD parameters $\boldsymbol{\Phi}$ include $N$ triples $\{(w_j,\boldsymbol{\mu}_j,\boldsymbol{\Sigma}_j)\}_{j=1}^N$ of the weights, means and covariance matrices. The weights $\mathbf{w}=[w_1,w_2,\ldots,w_N]^\top\in \mathcal{W}$ are constrained to the simplex
\begin{equation}\label{eq:weight_domain}
\mathcal{W}=\left\{\mathbf{w}\in\mathbb{R}^{N}\mid w_j>0,\sum_{j=1}^{N}w_j=1\right\},
\end{equation}
which ensures that Eq.~(\ref{eq:gmd_density}) is a valid PDF satisfying both nonnegativity and normalization. $\bm{\mu}_j=[\mu_{j1},\cdots,\mu_{jM}]^\top\in\mathbb{R}^{M}$ is the mean vector and $\bm{\Sigma}_j\in\mathcal{S}^M_{++}\subset\mathbb{R}^{M\times M}$ is the positive definite covariance matrix. The PDF of the $j$-th Gaussian component is
\begin{equation}
\mathcal{N}\left(\boldsymbol{\Theta};\boldsymbol{\mu}_j,\boldsymbol{\Sigma}_j\right)
=\frac{1}{(2\pi)^{M/2}|\boldsymbol{\Sigma}_j|^{1/2}}\exp
\left[-\frac{1}{2}(\boldsymbol{\Theta}-\boldsymbol{\mu}_j)^\top\boldsymbol{\Sigma}_j^{-1}(\boldsymbol{\Theta}-\boldsymbol{\mu}_j)\right].\label{eq:gaussian_density}
\end{equation}
The off-diagonal elements of $\boldsymbol{\Sigma}_j$ explicitly characterize the interdependencies among all parameters of the deterministic system and noises.

The parameter domain of the GMD in Eq.~(\ref{eq:gmd_density}) is a constrained set
\begin{equation}
\mathcal{Q}=\mathcal{W}\times\mathbb{R}^{MN}\times(\mathcal{S}^M_{++})^N.
\end{equation}
However, the normalization of the weights and positive definiteness of the covariance matrices are difficult algebraic properties for neural networks to directly obey. Therefore, we now build a surjection $f_\text{SUR}(\mathbf{y})=\boldsymbol{\Phi}$
\begin{equation}\label{eq:surjection}
f_\text{SUR}:\mathbb{R}^{[1+M(M+3)/2]N}\rightarrow\mathcal{Q},
\end{equation}
which transforms between feasible GMD parameters $\boldsymbol{\Phi}$ and an unconstrained vector $\mathbf{y}$. This design hard-encodes all constraints by allowing the neural network to output only the free vector $\mathbf{y}$, which significantly simplifies network construction and ensures that the constraints are inherently satisfied during optimization.

In this surjection, the first $N$ elements of $\mathbf{y}$ are mapped onto $\mathcal{W}$ by the Softmax function
\begin{equation}
\mathbf{w}=\mathrm{Softmax}(y_1,\cdots,y_N)=\left[\frac{\mathrm{e}^{y_1}}{\sum_{i=1}^N \mathrm{e}^{y_i}},\cdots,\frac{\mathrm{e}^{y_N}}{\sum_{i=1}^N \mathrm{e}^{y_i}}\right]^\top.
\end{equation}
To avoid numerical overflow in the exponential terms, we compute the maximum logit $y_\text{max}=\max_{1\leq i\leq N}y_i$ and use the numerically stable equivalent
\begin{equation}
\mathbf{w}=\mathrm{Softmax}(y_1-y_\text{max},\cdots,y_N-y_\text{max}),
\end{equation}
where all exponents are non-positive. The middle $MN$ elements of $\mathbf{y}$ are directly taken as the $N$ $M$-D means $\{\boldsymbol{\mu}_j\}_{j=1}^N$. 

To get rid of the constraints of the covariance matrix, we further utilize Cholesky decomposition to represent the inverse covariance matrix $\boldsymbol{\Sigma}_j^{-1}$,
\begin{equation}
\boldsymbol{\Sigma}_j^{-1}=\mathbf{U}_j^\top\mathbf{U}_j,\label{eq:precision_cholesky_L}
\end{equation}
where the upper triangular matrix $\mathbf{U}_j$ is parameterized by
\begin{equation}
\mathbf{U}_j=
\begin{bmatrix}
u_{j,11} & u_{j,12} & \cdots & u_{j,1M}\\
0        & u_{j,22} & \cdots & u_{j,2M}\\
\vdots   & \vdots   & \ddots & \vdots\\
0        & 0        & \cdots & u_{j,MM}
\end{bmatrix}.
\label{eq:upper_L}
\end{equation}
The $(M-1)M/2$ upper triangular off-diagonal elements $u_{j,rs},(r<s)$ are real. The $M$ diagonal elements $\{u_{j,ii}\}_{i=1}^M$ are positive. The Softplus function is further applied to replace these positive values by $M$ real, unconstrained quantities $\{\gamma_{j,ii}\}_{i=1}^M$
\begin{equation}\label{eq:diag_softplus}
u_{j,kk}=\operatorname{Softplus}(\gamma_{j,kk})=\ln\left(1+\exp(\gamma_{j,kk})\right), \quad k=1,2,\cdots,M.
\end{equation}
Therefore, a covariance matrix is parameterized by an $M(M+1)/2$-D real vector that represents the diagonal and off-diagonal elements.

By putting the four parts together, the GMD defined in Eqs.~(\ref{eq:gmd_density})-(\ref{eq:gaussian_density}) is parameterized by an unconstrained real vector $\mathbf{y}=[y_1,\cdots,y_{[1+M(M+3)/2]N}]^\top$. The surjection $f_\text{SUR}$ in Eq.~(\ref{eq:surjection}) that recovers the GMD parameters can be precisely expressed by
\begin{equation}\label{eq:surjection_function}
\left\{\begin{array}{ll}
w_j=\text{e}^{y_j}/(\sum_{k=1}^{N}\text{e}^{y_k}),&1\leq j\leq N,\\
\mu_{jk}=y_{N+(j-1)M+k},&1\leq j\leq N,1\leq k\leq M,\\
\gamma_{j,kk}=y_{N+MN+(j-1)M+k},&1\leq j\leq N,1\leq k\leq M,\\
u_{j,kl}=y_{N+2MN+(j-1)M(M-1)/2+r(k,l)},&1\leq j\leq N,1\leq k< l\leq M,
\end{array}\right.
\end{equation}
where $r(k,l)=\sum_{i=1}^{k-1}(M-i)+(l-k)$ for $1\leq k<l\leq M$ is the row-major position index of the upper-triangular entry $(k, l)$ in the vectorization. Eq.~(\ref{eq:surjection_function}) implies that the real vector $\mathbf{y}$ concatenates the weights, means, and the diagonal and off-diagonal elements of the upper triangular matrices of the $N$ Gaussian components. The four factors are represented by $N$, $NM$, $NM$, and $N[M(M-1)/2]$ real elements, respectively.

\begin{theorem}\label{the:1}
The mapping $f_\mathrm{SUR}:\mathbb{R}^{[1+M(M+3)/2]N}\to\mathcal{Q}$ defined in Eq.~(\ref{eq:surjection_function}) is surjective.
\end{theorem}

\begin{proof}
For any feasible GMD parameters $\boldsymbol{\Phi}=\{(w_j,\boldsymbol{\mu}_j,\boldsymbol{\Sigma}_j)\}_{j=1}^N\in\mathcal{Q}$, we construct a preimage $\mathbf{y}$ as follows.

For the weights, since $w_j>0$ and $\sum_{j=1}^N w_j=1$, choose $y_j=\ln w_j$ for $j=1,\dots,N$. Then $\mathbf{w}=\operatorname{Softmax}(y_1,\cdots,y_N)$, so the weights are attainable. For the means, the mapping is the identity on $\mathbb{R}^{MN}$. We directly set the entries of $\boldsymbol{\mu}_j$ back to the corresponding entries of $\mathbf{y}$.

For each covariance matrix $\boldsymbol{\Sigma}_j\in\mathcal{S}^M_{++}$, its inverse $\boldsymbol{\Sigma}_j^{-1}$ is symmetric positive definite and admits a unique Cholesky decomposition $\boldsymbol{\Sigma}_j^{-1}=\mathbf{U}_j^\top\mathbf{U}_j$, where $\mathbf{U}_j$ is an upper triangular matrix with strictly positive diagonal entries $u_{j,kk}>0$. The off-diagonal entries $u_{j,kl}$ ($k<l$) are real. Since the Softplus function in Eq.~(\ref{eq:diag_softplus}) is bijective, there exists a unique real number $\gamma_{j,kk}=\ln(\exp(u_{j,kk})-1)$ such that $u_{j,kk}=\operatorname{Softplus}(\gamma_{j,kk})$. We assign the off-diagonal entries directly as the corresponding components of $\mathbf{y}$, and assign $\gamma_{j,kk}$ to the diagonal positions as specified in Eq.~(\ref{eq:surjection_function}).

Concatenating the preimages of the weights, means, and all Cholesky factors yields a vector $\mathbf{y}\in\mathbb{R}^{[1+M(M+3)/2]N}$ such that $f_\mathrm{SUR}(\mathbf{y})=\boldsymbol{\Phi}$. Since $\boldsymbol{\Phi}$ was chosen arbitrarily, $f_\mathrm{SUR}$ is surjective.
\end{proof}

Substituting Eq.~(\ref{eq:precision_cholesky_L}) into Eq.~(\ref{eq:gaussian_density}), the equivalent PDF of the GMD is
\begin{equation}\label{eq:gmd_L_form}
q(\boldsymbol{\Theta};\boldsymbol{\Phi})=q(\boldsymbol{\Theta};f_\text{SUR}(\mathbf{y}))=\sum_{j=1}^{N}w_j\cdot\frac{|\mathbf{U}_j|}{(2\pi)^{M/2}}\mathrm{exp}\{-\frac{1}{2}\left\|\mathbf{U}_j(\boldsymbol{\Theta}-\boldsymbol{\mu}_j)\right\|_2^2\},
\end{equation}
where the determinant $|\mathbf{U}_j|=\prod_{k=1}^M u_{j,kk}=\prod_{k=1}^M \text{Softplus}(\gamma_{j,kk})$. With all GMD constraints hard-encoded in the surjection, the network only needs to map the trajectory to the free vector. Theorem~\ref{the:1} guarantees that, provided $\mathbf{y}$ can be freely generated, our method can theoretically recover any $N$-component GMD without loss of generality.

We define the mean of the GMD $q(\boldsymbol{\Theta};\boldsymbol{\Phi})$ as a point estimate:
\begin{equation}\label{eq:gmd_mean}
\hat{\boldsymbol{\Theta}}=[\hat{\theta}_1,\cdots,\hat{\theta}_M]^\top=\sum_{j=1}^N w_j\boldsymbol{\mu}_j.
\end{equation}
It is analogous to the outputs of earlier PENN frameworks~\cite{Wang2022Neural,Wang2025Noise}, providing a direct baseline for comparison. As will be shown in Sec.~\ref{sec:numer_ex}, the point estimate offers useful diagnostic signals of where estimation fails, while the parameter uncertainties it misses are fully captured by the GMD.

Moreover, the total covariance matrix of the GMD is computed by
\begin{equation}\label{eq:gmd_total_cov}
\boldsymbol{\Sigma}_{\mathrm{GMD}} = \sum_{j=1}^{N} w_j \left[\boldsymbol{\Sigma}_j + (\boldsymbol{\mu}_j - \hat{\boldsymbol{\Theta}})(\boldsymbol{\mu}_j - \hat{\boldsymbol{\Theta}})^\top \right].
\end{equation}
Its diagonal entries $\hat{\sigma}_{\theta_k}^2 = [\boldsymbol{\Sigma}_{\mathrm{GMD}}]_{kk}$ for $k=1,\cdots,M$ provide the marginal estimation variances for individual parameters, while the off-diagonal entries encode the correlations among parameter pairs. The statistical calibration of these diagonal variances will be systematically examined via a standardized residual analysis in Sec.~\ref{sec:dvdp}, and the off-diagonal correlations will be visualized and interpreted in Secs.~\ref{sec:dvdp} and~\ref{sec:airfoil}.

\subsection{Encoder and decoder architectures}
\label{sec:nn}
We now detail the design of the encoder $f_\text{ENC}$ and the decoder $f_\text{DEC}$. The encoder compresses the observed trajectory into a compact representation $\mathbf{e}\in \mathbb{R}^{D_\mathrm{EMB}}$, and the decoder refines this representation into a free vector $\mathbf{y}\in\mathbb{R}^{(1+M(M+3)/2)N}$, which is then mapped to feasible GMD parameters via the surjection $f_\text{SUR}$.

Given a sample trajectory $\mathbf{z}_{1:n}$ of length $n$, the encoder first utilizes a multi-layer LSTM network to recursively process the sample sequence, extracting $n$ local features $\mathbf{h}_{1:n}$,
\begin{equation}
\{\mathbf{h}_i\}_{i=1}^{n}=f_{\mathrm{LSTM}}(\mathbf{z}_{1:n}), \quad \mathbf{h}_i\in\mathbb{R}^{D_{\mathrm{EMB}}},
\label{eq:lstm_feature}
\end{equation}
where $D_{\mathrm{EMB}}$ denotes the dimension of the temporal feature vector extracted by the LSTM at each time step. The gated structure of the LSTM enables it to retain historical information during the recursive process while suppressing irrelevant disturbances in the observed trajectory. Therefore, it is suited to processing time series data of variable lengths and with partial observations.

The information relevant to parameter identification is largely local. The drift term of the stochastic system controls the states temporally and the autocorrelation functions of noises often decay along the time. Therefore, the local information in the trajectory carries substantial information about the underlying parameters. On the other hand, the LSTM is better at extracting short-term memory than maintaining very long history~\cite{pmlr-v119-zhao20c,chien2021slower}. Therefore, we do not rely solely on the LSTM to extract global information from the trajectory. Instead, in the second stage of the encoder, the local LSTM features from all time steps are averaged to obtain the trajectory embedding $\mathbf{e}$,
\begin{equation}\label{eq:feature_fusion}
\mathbf{e}=\frac{1}{n}\sum_{i=1}^{n}\mathbf{h}_i  \in\mathbb{R}^{D_{\mathrm{EMB}}}.
\end{equation}
The average pooling operator filters out inconsistent patterns, enhancing robustness. This design allows the LSTM to focus on extracting local temporal features without being burdened by long-range dependencies. The trajectory embedding $\mathbf{e}$ is assumed to include all information about system parameters governing the trajectory, including their estimates and uncertainties. The decoder is then designed to recover the GMD representation $\mathbf{y}$ from the embedding.

The decoder consists of a ResNet comprising $R$ residual blocks followed by a final outputting linear layer $l_{3R+1}$,
\begin{equation}
\mathbf{y}=f_\text{DEC}(\mathbf{e})=l_{3R+1}(B_R(B_{R-1}(\cdots B_2(B_1(\mathbf{e}))))).
\end{equation}
The $i$-th residual block $B_i$ is the vector sum of three linear layers followed by ELU activation functions~\cite{clevert2016fast} and a residual connection $s_i$ that sends the input directly to the output side,
\begin{equation}
B_i(\mathbf{u})=f_\text{ELU}(l_{3i}(f_\text{ELU}(l_{3i-1}(f_\text{ELU}(l_{3i-2}(\mathbf{u}))))))+s_i(\mathbf{u}).
\end{equation}
The linear layer $l_k$ is defined by an affine mapping
\begin{equation}
l_k(\mathbf{u})=\mathbf{W}_k\mathbf{u}+\mathbf{b}_k.
\end{equation}
For simplicity, we use the same number of neurons $N_\text{NEU}$ in all the linear layers except the first linear layer $l_1$ in the first residual block, where the input dimension is $D_\text{EMB}$, and the outputting linear layer after the last residual block, where the output dimension is $[1+M(M+3)/2]N$. The residual connection is defined by
\begin{equation}
s_i(\mathbf{u})=\left\{\begin{array}{ll}
l_0(\mathbf{u}), & i=1,\\
\mathbf{u}, & i>1.
\end{array}\right.
\end{equation}
Only the residual connection in the first residual block is a linear layer $l_0$ to adjust the length of the input vector from $D_\text{EMB}$ to $N_\text{NEU}$ to align with the output side. All the other residual connections in the last $R-1$ residual blocks are simply the identity function.

It is worth noting that we only utilize a single ResNet to construct the mapping from trajectory embedding to the weights, means, and the elements that determine the covariance matrices. The separation is achieved by splitting the output $\mathbf{y}$ of the ResNet. This strategy is simple and effective. In the preliminary study we used several sub-networks for calculating the weights, means and covariance matrices and found no improvement in accuracy. As these quantities are correlated, a single neural network can fully utilize the shared intermediate neuron weights, benefiting information flow.

\subsection{Loss function}
\label{sec:loss}
For a sample trajectory $\mathbf{Z} = \mathbf{z}_{1:L}$ with the ground-truth parameters $\bm{\Theta}^*$, the best GMD fit $\bm{\Phi} = \text{PENN}(\mathbf{Z})$ should minimize the NLL of $\bm{\Theta}^*$. By using Eq.~(\ref{eq:gmd_L_form}), the NLL is
\begin{equation}
-\ln q(\bm{\Theta}^*;\bm{\Phi})
= -\ln\left[\sum_{j=1}^{N}\frac{w_j|\mathbf{U}_j|}{(2\pi)^{M/2}}
\exp\left(-\frac{1}{2}\left\|\mathbf{U}_j(\bm{\Theta}^*-\bm{\mu}_j)\right\|_2^2\right)\right].
\end{equation}
By introducing
\begin{equation}
u_j = \ln w_j + \ln|\mathbf{U}_j| - \frac{1}{2}\left\|\mathbf{U}_j(\bm{\Theta}^*-\bm{\mu}_j)\right\|_2^2 - \frac{M}{2}\ln(2\pi),
\end{equation}
the NLL can be written as
\begin{equation}\label{eq:ori_nll}
-\ln q(\bm{\Theta}^*;\bm{\Phi}) = -\ln\sum_{j=1}^N e^{u_j}.
\end{equation}

Direct evaluation of the sum of exponentials in Eq.~(\ref{eq:ori_nll}) is prone to numerical overflow when one term $u_j$ dominates the others. To circumvent this issue, we apply the Log-Sum-Exp trick~\cite{Blanchard2021Accurately}, which yields the following numerically stable expression for the NLL of a single trajectory:
\begin{equation}\label{eq:nll_single}
L(\mathbf{Z},\bm{\Theta}^*) = -u_{\max} - \ln\sum_{j=1}^N e^{u_j - u_{\max}}, \quad u_{\max} = \max_{1\leq j\leq N} u_j.
\end{equation}
Since $u_j - u_{\max} \le 0$ for all $j$, all exponentials are bounded between 0 and 1, ensuring numerical stability without risk of overflow.

For a training batch of $N_\text{batch}$ trajectories $\{(\mathbf{Z}_i,\bm{\Theta}_i^*)\}_{i=1}^B$, the overall loss is the average NLL:
\begin{equation}\label{eq:batch_loss}
\mathcal{L}_{\text{batch}} = \frac{1}{N_\text{batch}}\sum_{i=1}^{N_\text{batch}} L(\mathbf{Z}_i,\bm{\Theta}_i^*).
\end{equation}
As the loss function is parameter-free and all the constraints of the GMD are hard-encoded into the mapping architecture, the optimization process is very simple. Minimizing Eq.~(\ref{eq:batch_loss}) trains the PENN-GMD to map an arbitrary raw trajectory $\mathbf{Z}$ to the mixture weights, means, and full covariance matrices of the GMD $q(\boldsymbol{\Theta};\boldsymbol{\Phi})$, which approximates the likelihood distribution $p(\mathbf{Z}|\boldsymbol{\Theta})$. The log-sum-exp stabilization ensures that the training remains numerically stable even when the GMD components are highly separated.

\subsection{Network training}
\label{sec:training}
Training the proposed method only requires the system equations to be known, so that trajectories can be simulated with parameters sampled from the domain $\mathcal{P}$. The PENN-GMD automatically learns the statistical dependencies among the system parameters and process noises directly from the partially observed training data.

All the tunable weights of the PENN are contained in the $3R+2$ linear layers $\{l_i\}_{i=0}^{3R+1}$ of the encoder $f_\text{ENC}$ and decoder $f_\text{DEC}$. Algorithm~\ref{alg:training} details the training process of the PENN-GMD. We use the ADAM~\cite{kingma2017adam} optimizer with a learning rate of 0.0002 to train the encoder and decoder.

\begin{algorithm}[htbp]
\caption{Training of the PENN-GMD framework}
\label{alg:training}
\begin{algorithmic}[1]
\Require
  Training dataset $\mathcal{D} = \{(\mathbf{Z}_j, \boldsymbol{\Theta}_j^*)\}_{j=1}^{N_{\text{train}}}$, batch size $N_\text{batch}$, number of epochs $N_\text{epoch}$.
\Ensure Network weights of $f_{\text{ENC}}$ and $f_{\text{DEC}}$.
\State Initialize all network weights randomly.
\For{$i=1$ \textbf{to} $N_\text{epoch}$}
  \State Shuffle the training dataset $\mathcal{D}$ and split it into mini-batches of size $N_\text{batch}$.
  \For{each mini-batch $\{(\mathbf{Z}_{j}, \boldsymbol{\Theta}_j^*)\}_{j=1}^{N_\text{batch}}$}
    \State Perform forward pass
          $\boldsymbol{\Phi}_j = \text{PENN}(\mathbf{Z}_j) = f_{\text{SUR}}\bigl(f_{\text{DEC}}(f_{\text{ENC}}(\mathbf{Z}_j))\bigr), \quad j=1,\dots,N_\text{batch}.$
    \State Compute the batch NLL loss using Eq.~(\ref{eq:batch_loss}).
    \State Perform backpropagation and update network parameters using the ADAM optimizer.
  \EndFor
\EndFor
\end{algorithmic}
\end{algorithm}

\section{Numerical experiments}
\label{sec:numer_ex}
The proposed method is tested on a linear system with known analytical likelihood and four paradigmatic stochastic systems with mixed process noises, covering a broad range of challenging practical scenarios. All the PENN-GMDs have the same architecture. The LSTM in the encoder has four layers and each layer includes $D_\text{EMB}=50$ neurons, which is also the dimensionality of the trajectory embedding $\mathbf{e}$. The ResNet in the decoder has 6 residual blocks and each linear layer has $N_\text{NEU}=100$ neurons. The GMD output has $N=10$ components. The differences among these PENN-GMDs are that their input dimensions are aligned with the dimensionality of the measurements $\mathbf{z}$, and the dimensionality of the output GMD is in accord with the respective system parameters.

A key insight of this work is that the point estimate in Eq.~(\ref{eq:gmd_mean}), i.e., the mean of the GMD, already serve as a diagnostic tool. The distribution of their errors reveals where and how parameter estimation fails, and points to the presence of parameter couplings and non-identifiability. However, the point estimate alone cannot quantify these structures. The GMD extension then provides the precise, closed-form characterization of these couplings and uncertainties, confirming and refining the patterns anticipated by the point-estimate diagnostics. We therefore structure each numerical example by first examining the diagnostic information provided by the point estimates, and then demonstrating how the GMD output reveals the full coupling and uncertainty structure that underlies these diagnostic patterns.

\subsection{Standard Ornstein-Uhlenbeck process}
\label{sec:soup}
We first investigate a standard Ornstein-Uhlenbeck process (SOUP) driven by a white Gaussian noise (WGN)
\begin{equation}\label{eq:soup}
\dot{x}(t) = -\frac{1}{\tau} x(t) + \sqrt{\frac{2D}{\tau}}\dot{B}(t),
\end{equation}
with correlation time $\tau$ and noise intensity $D$ in the parameter domain
\begin{equation}
\mathcal{P}_\text{SOUP}=\{\boldsymbol{\Theta}=[\lg\tau, D]^\top|\lg\tau\in[-0.5, 1.5];\;D\in[0.01, 1]\}.
\end{equation}
In Eq.~(\ref{eq:soup}), the overhead dots indicate derivatives with the time $t$. The first term in the right hand side governs the mean reversion of the SOUP, where a large correlation time $\tau$ leads to slower reversion and more persistent trajectories. We parameterize the common logarithm $\lg\tau$ of $\tau$ to cover different magnitudes of reversions. In the second term, $B(t)$ is a standard Brownian motion. Its formal derivative $\dot{B}(t)$ is the standard WGN with the autocorrelation function $\langle\dot{B}(t)\dot{B}(t+s)\rangle=\delta(s)$. The SOUP has stationary variance $\mathrm{Var}[x(t)] = D$ and autocorrelation function $\langle x(t)x(t+s)\rangle = D \mathrm{e}^{-s/\tau}$. 

A key advantage of the SOUP is that the likelihood $p(x_{1:n}|\boldsymbol{\Theta})$ can be computed analytically via the exact discrete-time representation. For equally spaced observations with sampling time $\Delta t$, the process obeys the AR(1) form~\cite{Gardiner2009Stochastic}
\begin{equation}
x_{i+1} = a x_i + \varepsilon_i, \quad a = \mathrm{e}^{-\Delta t/\tau}, \quad \varepsilon_i \sim \mathcal{N}(0,\sigma^2),
\end{equation}
with the innovation variance 
\begin{equation}\label{eq:ar1_var}
\sigma^2 = D(1 - \mathrm{e}^{-2\Delta t/\tau})\approx2\Delta t \cdot \frac{D}{\tau}.
\end{equation}
Assuming the process is stationary, i.e., $x_1 \sim \mathcal{N}(0, D)$, the likelihood function is
\begin{equation}\label{eq:soup_likelihood}
p(x_{1:n}|\boldsymbol{\Theta})
= \frac{1}{\sqrt{2\pi D}}
\exp\left(-\frac{x_1^2}{2D}\right)
\prod_{i=1}^{n-1}
\frac{1}{\sqrt{2\pi\sigma^2}}
\exp\left(-\frac{(x_{i+1} - a x_i)^2}{2\sigma^2}\right).
\end{equation}
This analytical likelihood serves as a ground truth to validate the GMD estimates produced by the PENN-GMD. Moreover, a key coupling is revealed through the innovation variance in Eq.~(\ref{eq:ar1_var}) as $\tau$ and $D$ act as complementary drivers of the observed variability. A larger $\tau$ smooths the process and must be offset by a larger $D$ to preserve the same fluctuation level. This structural relationship is an inherent property of the OUP, making it an ideal test case for the GMD covariance structure.

\begin{figure}[!htb]
\center{\includegraphics[width=1\textwidth]
{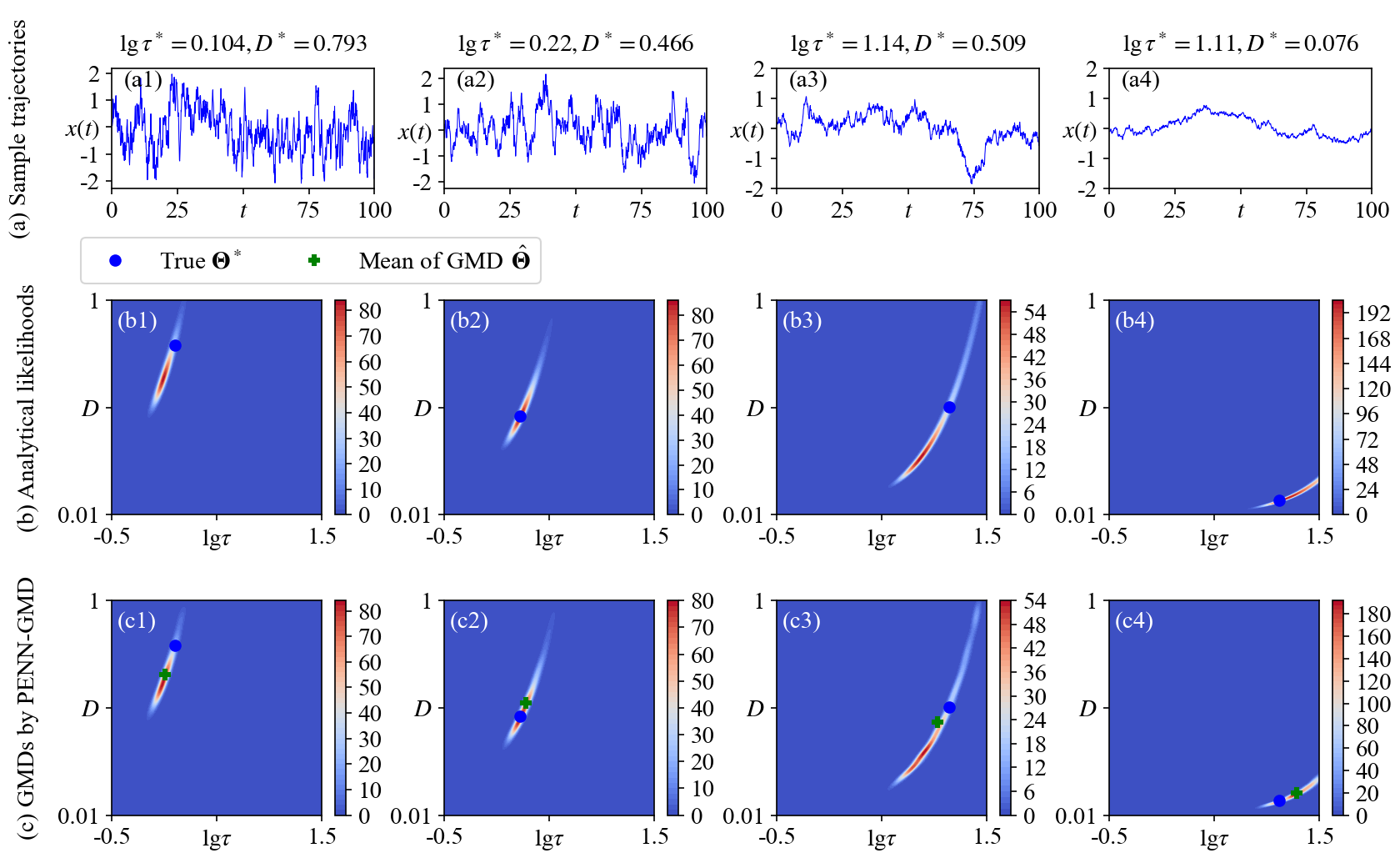}}
\caption{\label{fig:soup_estimates}A comparison of the proposed GMD estimates and the analytical likelihood functions on the SOUP in Eq.~(\ref{eq:soup}). (a) Four sample trajectories for parameter identification. (b) The analytical likelihood distributions of the four sample trajectories by Eq.~(\ref{eq:soup_likelihood}). (c) The GMDs obtained by PENN-GMD. The distributions in (b) and (c) are normalized to have integral equal to one.}
\end{figure}

Fig.~\ref{fig:soup_estimates}(a) shows four sample trajectories with different $(\tau, D)$ pairs. Each trajectory is simulated using the Euler-Maruyama method with integration step $\delta t = 0.01$ and sampling time $\Delta t=0.1$, i.e., recording measurements every 10 integration steps. For each trajectory, the samples within $t_0 = 500\Delta t = 50$ are discarded to remove transient effects and then $L = 1000$ time steps of full states ($\mathbf{z} = [x]$) are recorded. Fig.~\ref{fig:soup_estimates}(b) shows the analytical likelihood functions for the same four trajectories by using Eq.~(\ref{eq:soup_likelihood}), confirming the positive correlation between $\lg\tau$ and $D$. It is worth noting that even in this simple example of unimodal likelihoods, the true parameters $\boldsymbol{\Theta}^*$ are often not located exactly at the mode of the sample-induced likelihood distribution, such as Figs.~\ref{fig:soup_estimates}(b1) and (b3).

To construct the proposed estimator, we simulate $N_\text{train}=300,000$ trajectories of Eq.~(\ref{eq:soup}) as the training dataset, where the parameters of each trajectory are uniformly sampled in $\mathcal{P}_\text{SOUP}$. A PENN-GMD is trained for $N_\text{epoch}=1200$ epochs, after which the loss curve on an independent validation set of 20,000 trajectories has converged. The batch size $N_\text{batch}$ is 1800 such that the memory of a GeForce RTX 4090 GPU is fully utilized without overflow. 

Fig.~\ref{fig:soup_estimates}(c) shows the GMD estimates produced by the PENN-GMD for the four trajectories in Fig.~\ref{fig:soup_estimates}(a). The GMDs closely match the ground-truth likelihood functions in Fig.~\ref{fig:soup_estimates}(b), confirming that the PENN-GMD provides accurate data-driven approximations to the true likelihood distributions. Fig.~\ref{fig:soup_estimates}(c) also shows that the GMD means $\hat{\boldsymbol{\Theta}}$ defined in Eq.~(\ref{eq:gmd_mean}) lie close to the true parameters $\boldsymbol{\Theta}^*$, indicating that the point estimate of the PENN-GMD is also accurate.

\subsection{Ornstein-Uhlenbeck process with fGn and L{\'e}vy noises}
\label{sec:oup}
We now consider the OUP driven by fGn and L{\'e}vy noise
\begin{equation}\label{eq:ou_fgn_levy}
\dot{x}(t) = -\frac{1}{\tau} x(t) + \sqrt{\frac{2D_{\mathrm{FGN}}}{\tau}} \dot{W}_H(t) + \left(\frac{D_{\mathrm{L\acute{e}vy}}}{\tau}\right)^{1/\alpha} \dot{L}_\alpha(t),
\end{equation}
with 5 parameters $\boldsymbol{\Theta}=[\lg\tau, D_{\mathrm{FGN}}, H, D_\mathrm{L\acute{e}vy},\alpha]^\top$ in the parameter domain
\begin{equation}
\mathcal{P}_\text{OUP}=\{\boldsymbol{\Theta}|\lg\tau\in[-0.5, 1.5];\;D_{\mathrm{FGN}},D_\mathrm{L\acute{e}vy}\in[0, 1];\; H\in[0.05, 0.95];\;\alpha\in[1.4, 2]\}.
\end{equation}
In the right hand side of Eq.~(\ref{eq:ou_fgn_levy}), the second term represents the fGn, a correlated noise, with the intensity $D_{\mathrm{FGN}}$. $W_H(t)$ is the fractional Brownian motion (fBm) and the fGn $\dot{W}_H(t)$ is its formal derivative~\cite{Feng2023Deep}. The Hurst exponent $H \in (0,1)$ governs the long-range dependence of the fGn. $H>0.5$ induces persistent behavior, $H<0.5$ induces anti-persistent behavior, and $H=0.5$ recovers standard WGN. The third term of Eq.~(\ref{eq:ou_fgn_levy}) is the symmetric $\alpha$-stable L{\'e}vy noise, a noise model with both large jumps and small fluctuations. $L_\alpha(t)$ is L{\'e}vy motion and its formal derivative $\dot{L}_\alpha(t)$ is the L{\'e}vy noise~\cite{Wang2022Neural}. $D_{\mathrm{L\acute{e}vy}}$ is the intensity of the L{\'e}vy noise and $\alpha$ is the stability index, which controls the tail heaviness of the noise distribution. Smaller $\alpha$ yields heavier tails and more frequent large jumps, while $\alpha=2$ recovers Gaussian noise. The two noise sources are mutually independent.

This linear system serves as a challenging testbed for parameter estimation as the  mixed noises exhibit rich stochastic behaviors. The fGn introduces long-range correlations with power-law decaying autocorrelation, while the L{\'e}vy noise introduces intermittent large-amplitude jumps. Moreover, as the fGn is a correlated noise with long-range dependency, the system is no longer Markovian, posing difficulties in computing the likelihood function. The L{\'e}vy noise, on the other hand, lacks a closed-form PDF expression for general $\alpha$ except for several special cases such as the Cauchy case $\alpha=1$ and the Gaussian case $\alpha=2$, again making it difficult to compute the likelihood of measurements. Most importantly, the system contains a special degeneracy point at $(H, \alpha) = (0.5, 2)$, where both noises reduce to WGNs. At this point, the two noise sources become indistinguishable from the WGN, and the system in Eq.~(\ref{eq:ou_fgn_levy}) reduces to the SOUP
\begin{equation}\label{eq:ou_gauss}
\dot{x}(t) = -\frac{1}{\tau} x(t) + \sqrt{\frac{2(D_{\mathrm{FGN}}+D_{\mathrm{L\acute{e}vy}})}{\tau}}\dot{B}(t),
\end{equation}
where only the sum intensity $D_{\mathrm{FGN}}+D_{\mathrm{L\acute{e}vy}}$ enters the equation, rendering the individual parameters $D_{\mathrm{FGN}}$ and $D_{\mathrm{L\acute{e}vy}}$ non-identifiable. This provides an ideal scenario to test whether the PENN-GMD can capture such non-identifiability through multi-modal or broad distributions.

A natural concern when applying a Gaussian mixture to systems driven by L{\'e}vy noise is that the noise PDF itself is heavy-tailed. The likelihood $p(\mathbf{Z}|\boldsymbol{\Theta})$ may inherit such heavy tails, as it is obtained by evaluating the noise density at the observed increments as a function of $\boldsymbol{\Theta}$. The GMD with exponentially decayed distribution can not precisely fit a heavy-tailed distribution. However, in the present setting, the parameter domain $\mathcal{P}$ is compact, and we only require the GMD to approximate the main probability mass of the likelihood distribution. The finite Gaussian mixture therefore serves as a practically sufficient parametric surrogate for the likelihood over the parameter region of interest.

To train the PENN-GMD, we simulate $N_\text{train}=300,000$ trajectories of Eq.~(\ref{eq:ou_fgn_levy}) with parameters uniformly sampled in $\mathcal{P}_\text{OUP}$. The fGn is generated via the Davies-Harte method~\cite{Davies1987Tests}, the L{\'e}vy noise is generated using the Chambers-Mallows-Stuck (CMS) algorithm~\cite{Chambers1976Method}, and other settings are the same as Sec.~\ref{sec:soup}. The PENN-GMD is trained using the batch size $N_\text{batch}=1800$ for $N_\text{epoch}=2500$ epochs.

\begin{figure}[!htb]
\center{\includegraphics[width=1\textwidth]
{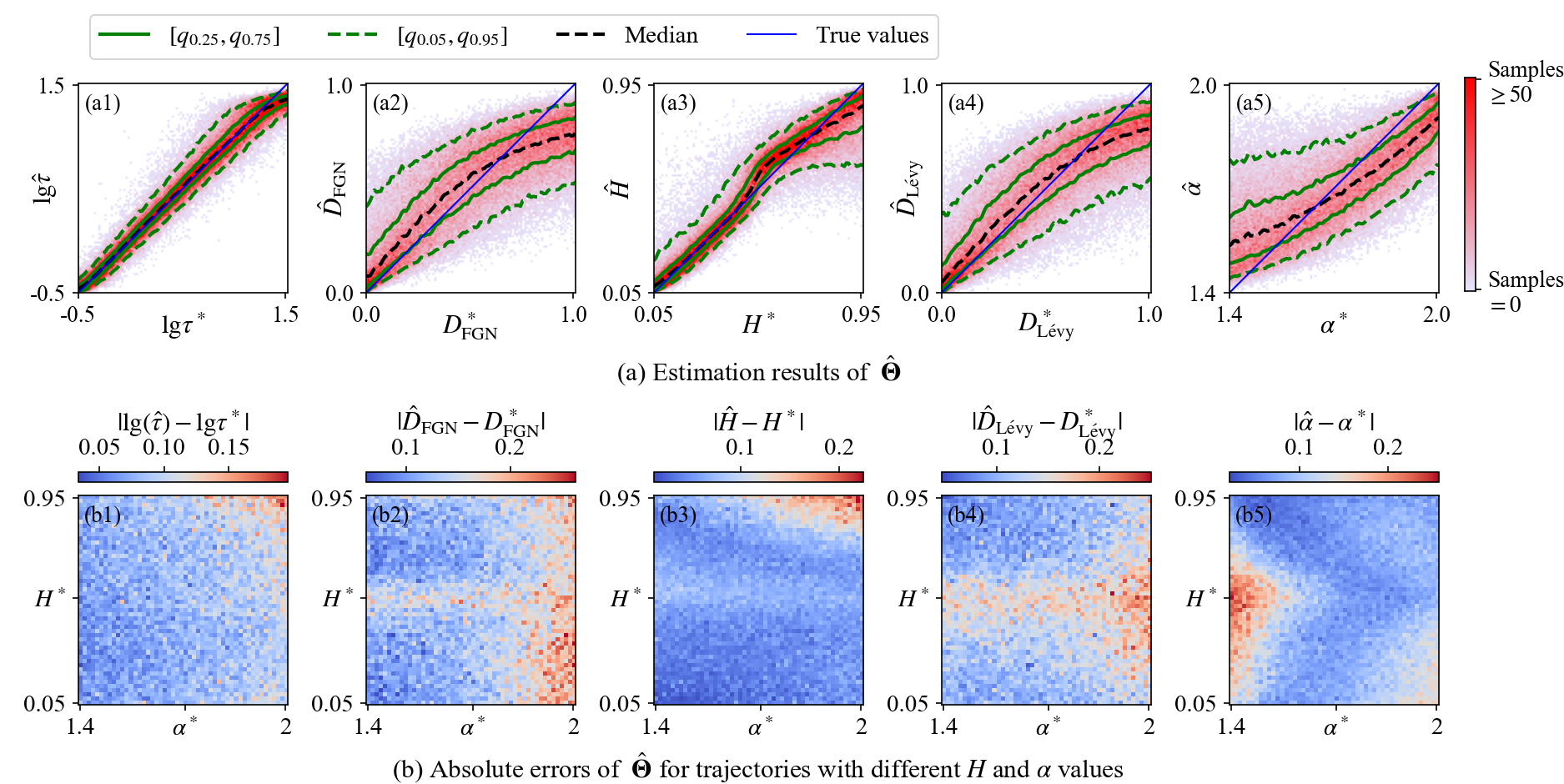}}
\caption{\label{fig:oup_estimates}Point estimates $\hat{\boldsymbol{\Theta}}$ by the PENN-GMD in Eq.~(\ref{eq:gmd_mean}) on 80,000 test trajectories of the OUP in Eq.~(\ref{eq:ou_fgn_levy}). (a) The histograms of point estimates $\hat{\boldsymbol{\Theta}}$ versus the true values $\boldsymbol{\Theta}^*$. The 5\%, 25\%, 50\%, 75\%, and 95\% quantiles are drawn. (b) The correlations between absolute estimation errors of the five parameters and the true $\alpha$ and $H$ of the trajectories.}
\end{figure}

The PENN-GMD is tested on 80,000 trajectories generated in a similar way as the training dataset. Fig.~\ref{fig:oup_estimates}(a) shows the 80,000 point estimates, i.e., the GMD means $\hat{\boldsymbol{\Theta}}$ in Eq.~(\ref{eq:gmd_mean}) versus the true parameters $\boldsymbol{\Theta}^*$. The perfect estimates are on the diagonals. It can be seen that in general, all parameters can be effectively estimated as the main distributions are aligned with the diagonals. Among the five parameters, the correlation time $\lg\tau$ is estimated the best as the distributions in Fig.~\ref{fig:oup_estimates}(a1) attain the least variances. The Hurst exponent $H$ can be accurately estimated when $H<0.6$, while the variances when $H>0.6$ are large. The uncertainties of the two noise intensities $D_{\mathrm{FGN}}$ and $D_{\mathrm{L\acute{e}vy}}$ and the stability index $\alpha$ are high, verifying their ambiguity and highlighting the fundamental weakness of the point estimation.

A key feature of the PENN-based point estimate is that we can quickly test a large number of trajectories and investigate the error distributions to infer the uncertainties among different parameter choices. Fig.~\ref{fig:oup_estimates}(b) visualizes the absolute errors of the point estimates $\hat{\boldsymbol{\Theta}}$ for the 80,000 trajectories as functions of the true $\alpha$ and $H$ values. The uniform histogram in Fig.~\ref{fig:oup_estimates}(b1) indicates that the error in $\ln\hat{\tau}$ is independent of the true stability index and Hurst exponent of the trajectories. In contrast, there is a T-shape region of high errors in Figs.~\ref{fig:oup_estimates}(b2) and (b4), consisting of a horizontal band along $H=0.5$ and a vertical band along $\alpha=2$. In such cases, the fGn and L{\'e}vy noise reduce to the same Gaussian noise such that the noise intensities $D_{\mathrm{FGN}}$ and $D_{\mathrm{L\acute{e}vy}}$ become non-identifiable, leading to large errors. Moreover, Fig.~\ref{fig:oup_estimates}(b5) indicates that when $H$ is close to 0.5 or $\alpha$ is close to 2, the error of $\hat{\alpha}$ is also high.

\begin{figure}[!htb]
\center{\includegraphics[width=1\textwidth]
{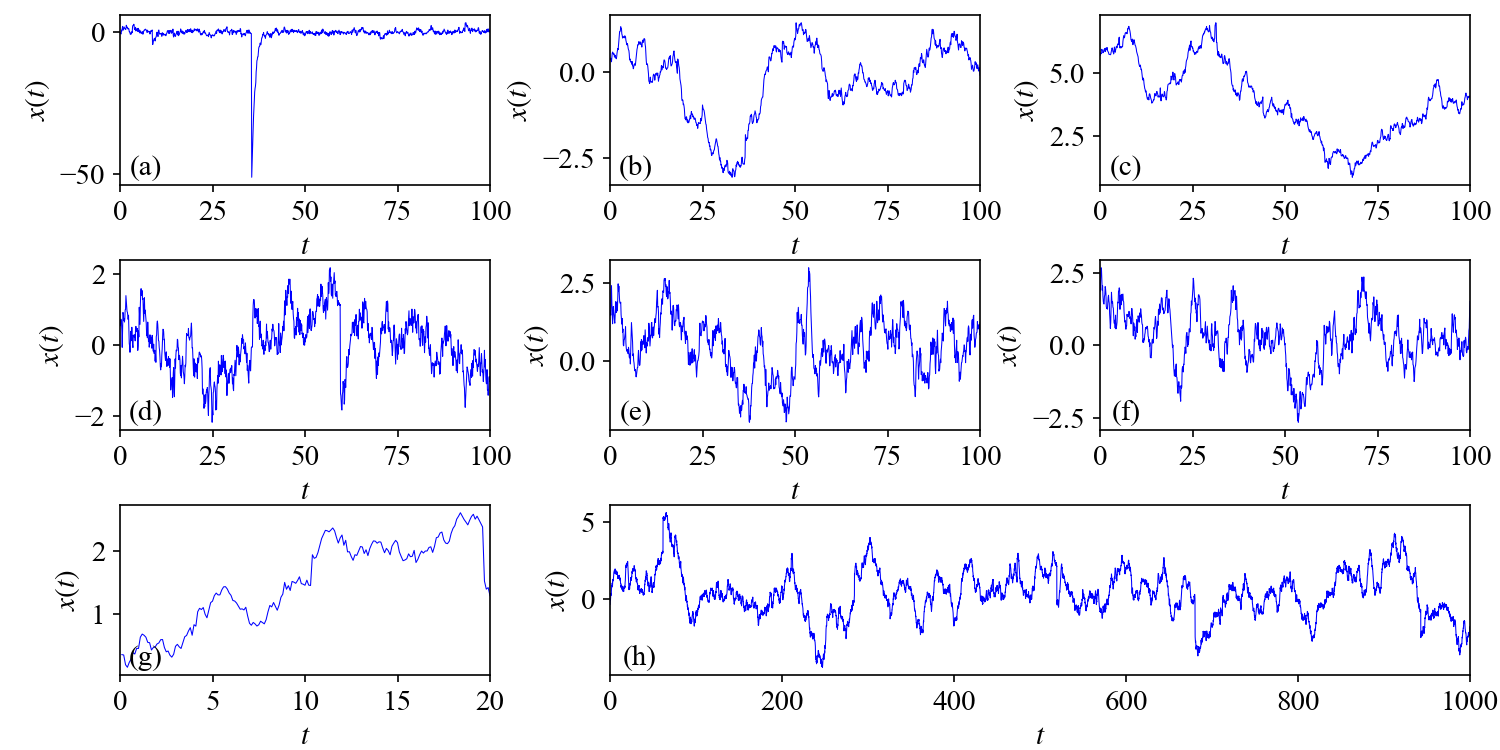}}
\caption{\label{fig:oup_trajectories}8 sample trajectories of the OUP in Eq.~(\ref{eq:ou_fgn_levy}). Their parameters and GMD estimates are detailed in Fig.~\ref{fig:oup_distri}.}
\end{figure}

\begin{figure}[!htb]
\center{\includegraphics[width=1\textwidth]
{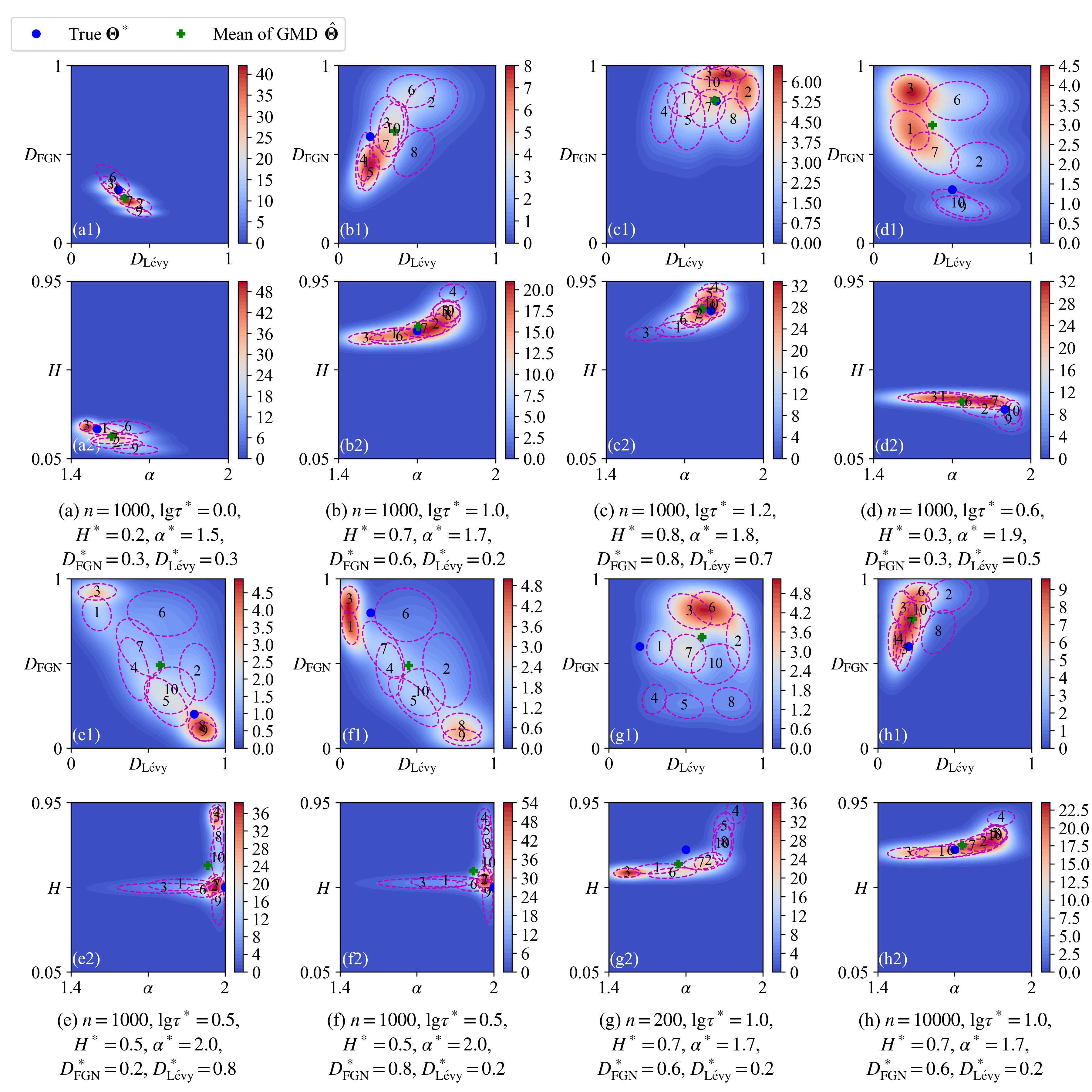}}
\caption{\label{fig:oup_distri}2-D marginal GMDs of $(D_{\mathrm{L\acute{e}vy}}, D_{\mathrm{FGN}})$ and $(\alpha, H)$ provided by the PENN-GMD on the 8 test trajectories in Fig.~\ref{fig:oup_trajectories} of the OUP in Eq.~(\ref{eq:ou_fgn_levy}). The 1-$\sigma$ ellipses of the Gaussian components with weights greater than 0.005 are drawn with labels.}
\end{figure}

To illustrate the ability of PENN-GMD to capture multi-modal likelihood distributions, we test 8 trajectories with different parameters and lengths and obtain the full GMD estimates by the PENN-GMD. The trajectories are plotted in Fig.~\ref{fig:oup_trajectories} and the true parameters are detailed in Fig.~\ref{fig:oup_distri}, where the 2-D marginal GMDs of noise intensities $(D_{\mathrm{L\acute{e}vy}},D_{\mathrm{FGN}})$ and noise characteristics $(\alpha, H)$ are drawn.

Figs.~\ref{fig:oup_distri}(a)-(c) show that when $H$ is distant from 0.5 and $\alpha$ is distant from 2, the marginal GMDs are concentrated and the point estimates $\hat{\boldsymbol{\Theta}}$ are close to the true values $\boldsymbol{\Theta}^*$. These patterns imply that the point estimates are already accurate. The PENN-GMD also successfully quantifies the uncertainty by producing localized GMDs, correctly indicating high confidences or low uncertainties about the estimates. In Fig.~\ref{fig:oup_distri}(d), although $\alpha=1.9$ is close to 2, PENN-GMD can still give correct and concentrated estimate of $H$. However, the variances of $\alpha$ and the two noise intensities are high, and the joint distribution of $D_{\mathrm{L\acute{e}vy}}$ and $D_{\mathrm{FGN}}$ shows strong negative correlation. These indicate that PENN-GMD correctly quantifies the uncertainties of the noise intensities as it is difficult to determine the noise source from the measurements with almost Gaussian noises.

The noise sources of the two trajectories in Figs.~\ref{fig:oup_distri}(e) and (f) are completely non-identifiable as $H=0.5$ and $\alpha=2$, where the system equation reduces to Eq.~(\ref{eq:ou_gauss}). The PENN-GMD remarkably produces all possible combinations. The high-probability regions of the noise intensities are approximately on the line $D_{\mathrm{L\acute{e}vy}}+D_{\mathrm{FGN}}=1$, the best estimate any estimator can achieve. Furthermore, the Gaussian components with high weights can be partitioned into three groups, precisely representing three probable situations. For example, the horizontal band in Fig.~\ref{fig:oup_distri}(e2) consists of the 1st and the 3rd components and corresponds to the upper-left high-probability region in Fig.~\ref{fig:oup_distri}(e1). It indicates that the fGn has Hurst exponent close to 0.5 and intensity around 1, while the L{\'e}vy noise has negligible intensity and its stability index is essentially non-identifiable. The vertical band consists of the 8th, 9th, 10th, and 5th components, indicating the dual situation where the L{\'e}vy noise has stability index close to 2 and intensity around 1, while the fGn has negligible intensity and its Hurst exponent is non-identifiable. The probability mass centered at $\alpha=2$ and $H=0.5$ in Fig.~\ref{fig:oup_distri}(e2) consists of the 2nd, 6th, and 7th components. In such situation, the probability mass of the noise intensities is scattered around the line $D_{\mathrm{L\acute{e}vy}}+D_{\mathrm{FGN}}=1$ as they are negatively correlated and cannot be individually determined. This partition of the likelihood distribution into three physically meaningful clusters clearly demonstrates that PENN-GMD reveals the entire manifold of parameter combinations compatible with the observed trajectory.

To show the capability of PENN-GMD on trajectories with different lengths, the trajectories in Figs.~\ref{fig:oup_trajectories}(g), (b) and (h) are generated with the same parameters with increasing lengths $n=200$, 1000, and 10,000 measurements. In Figs.~\ref{fig:oup_distri}(g), (b) and (h), the variances of the corresponding GMDs of longer trajectories are visibly smaller, confirming that the PENN-GMD effectively utilizes additional temporal information to reduce estimation uncertainty. This also aligns with the intuition that longer trajectories provide more information about the underlying noise statistics.

\subsection{Duffing-Van der Pol system with colored noises}
\label{sec:dvdp}
We next consider the second-order Duffing-Van der Pol system (DVDP)~\cite{Xu2011Stochastic}
\begin{equation}\label{eq:sys_dvdp}
\begin{split}
\dot{x} &= y,\\
\dot{y} &= (\varepsilon + \beta_1 x^2 - \beta_2 x^4) y - x - \beta_0 x^3 + \eta + x \xi,\\
\dot{\eta} &= -\frac{\eta}{\tau_1} + \frac{\sqrt{2D_1}}{\tau_1}\dot{B}_1(t),\\
\dot{\xi} &= -\frac{\xi}{\tau_2} + \frac{\sqrt{2D_2}}{\tau_2}\dot{B}_2(t),
\end{split}
\end{equation}
with 8 parameters $\boldsymbol{\Theta}=[\varepsilon, \beta_0, \beta_1, \beta_2, D_1, D_2, \tau_1, \tau_2]^\top$ in the parameter domain
\begin{equation}
\begin{split}
\mathcal{P}_\text{DVDP}=\{\boldsymbol{\Theta}|&\varepsilon\in[-0.3, 0.1];\;\beta_0\in[0, 0.2];\; \beta_1,\beta_2\in[0.2, 1];\\
&D_1,D_2\in[0.1, 0.5];\;\tau_1,\tau_2\in[0.5, 5]\}.
\end{split}
\end{equation}
Here, $\eta$ and $x\xi$ are additive and multiplicative colored noises, respectively. Both are generated by independent SOUPs with different correlation times and noise intensities, as shown in the third and fourth equations of Eq.~(\ref{eq:sys_dvdp}). $B_1(t)$ and $B_2(t)$ are two independent Brownian motions whose formal derivatives are two independent WGNs.

From the perspective of parameter identification, we can infer from Eq.~(\ref{eq:sys_dvdp}) that $\beta_1$ and $\beta_2$ are positively correlated, since increasing $\beta_1$ may partially compensate for an increase in $\beta_2$ in the term $(\beta_1 x^2 - \beta_2 x^4)y$. In the two SOUPs, the inference in Sec.~\ref{sec:soup} implies that $\tau_1$ and $\tau_2$ are positively correlated with $D_1$ and $D_2$, respectively. These parameter couplings, rooted in the system dynamics and statistical properties of the colored noises, can be captured by the full covariance matrices of the GMD. It is also interesting to evaluate whether the additive and multiplicative noises can be effectively identified without confusion.

We train a PENN-GMD on $N_\text{train}=500,000$ trajectories of Eq.~(\ref{eq:sys_dvdp}) using the Euler-Maruyama method with integration step $\delta t = 0.01$ and sampling time $\Delta t=0.1$. Each trajectory contains $L = 500$ time steps and the samples within $t_0 = 2000\Delta t = 200$ are removed to exclude transients. We only use the $x$ measurements, i.e., $\mathbf{z}=[x]$ and treat $y$, $\eta$, and $\xi$ as unobservable. The PENN-GMD is trained for $N_\text{epoch}=2000$ epochs with batch size $N_\text{batch}=1200$.

\begin{figure}[!htb]
\center{\includegraphics[width=0.8\textwidth]
{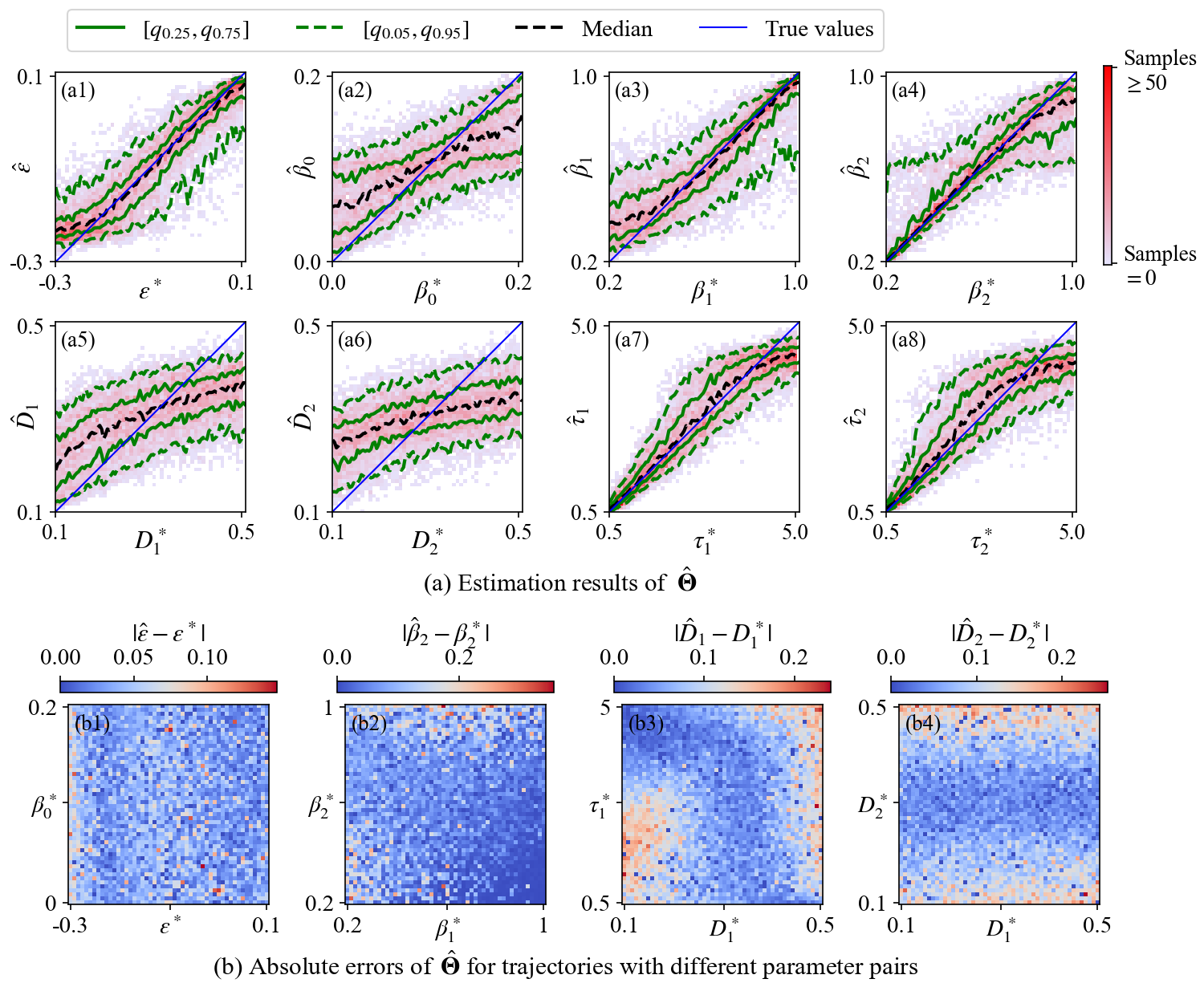}}
\caption{\label{fig:dvdp_estimates}Point estimates by the PENN-GMD on 10,000 test trajectories of the DVDP in Eq.~(\ref{eq:sys_dvdp}). (a) The histograms and quantiles of point estimates versus the true values. (b) The correlations between estimation errors and true parameters of the trajectories.}
\end{figure}

Fig.~\ref{fig:dvdp_estimates}(a) shows the point estimates $\hat{\boldsymbol{\Theta}}$ of the PENN-GMD on 10,000 test trajectories, which are generated in a similar way to the training dataset. The variances of the estimates are high for all parameters. Severe overestimation occurs in Figs.~\ref{fig:dvdp_estimates}(a1) to (a6) when the true parameter values are close to the lower bounds of the sampling ranges in $\mathcal{P}_\text{DVDP}$ while severe underestimation occurs for all the 8 parameters when the true parameter values are close to the upper bounds, showing that the point estimates are poor.

Fig.~\ref{fig:dvdp_estimates}(b) further details the correlations between the absolute errors of $\hat{\boldsymbol{\Theta}}$ and the true parameters across four subfigures. Fig.~\ref{fig:dvdp_estimates}(b1) shows that the error of $\hat{\varepsilon}$ is largely independent of the true $\varepsilon$ and $\beta_0$ values, implying they are uncorrelated. Figs.~\ref{fig:dvdp_estimates}(b2) and (b3) present complex error distributions, indicating that the error of $\hat{\beta}_2$ depends on the joint true $\beta_1$ and $\beta_2$ values, and the error of $\hat{D}_1$ depends on both $D_1$ and $\tau_1$. The errors of $\hat{D}_2$ in Fig.~\ref{fig:dvdp_estimates}(b4) show horizontal contour lines, implying that the estimate of $D_2$ is uncorrelated with $D_1$. Moreover, the errors are low only when the true $D_2$ values are near the center 0.3 of its sampling range $[0.1, 0.5]$, indicating that the point estimates severely shrink toward the center, which is a clear sign of unidentifiability. These observations suggest that point estimation is not suitable here, not only because the parameters are correlated, but also because the trajectory length of 500 samples is insufficient to determine accurate parameters for this second-order nonlinear system from partially observed measurements.

\begin{figure}[!htb]
\center{\includegraphics[width=1\textwidth]
{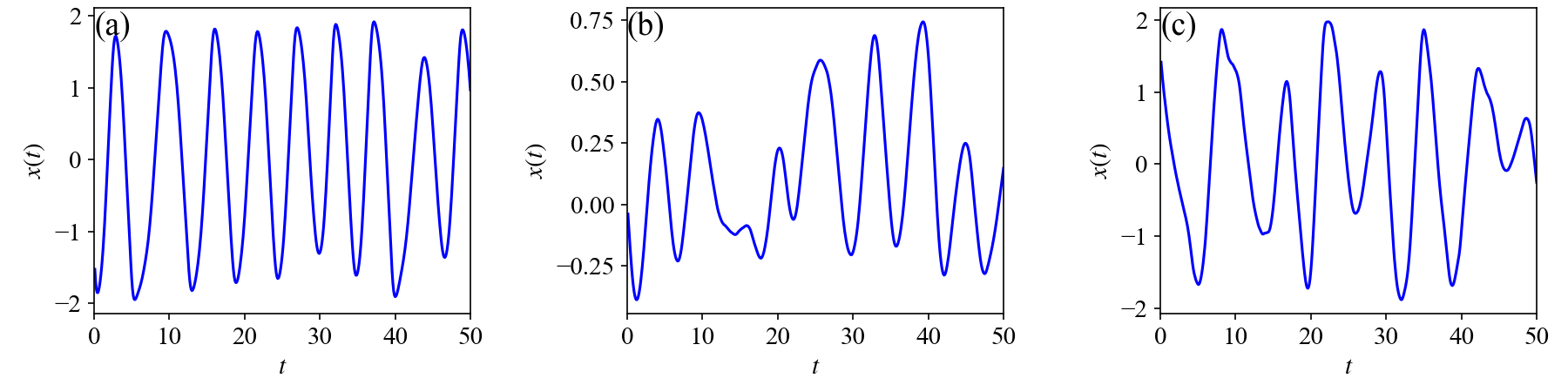}}
\caption{\label{fig:dvdp_trajectories}Three $x(t)$ trajectories of the DVDP system. The system parameters and the GMD estimates by the PENN-GMD are detailed in Fig.~\ref{fig:dvdp_distri}.}
\end{figure}

\begin{figure}[!htb]
\center{\includegraphics[width=1\textwidth]
{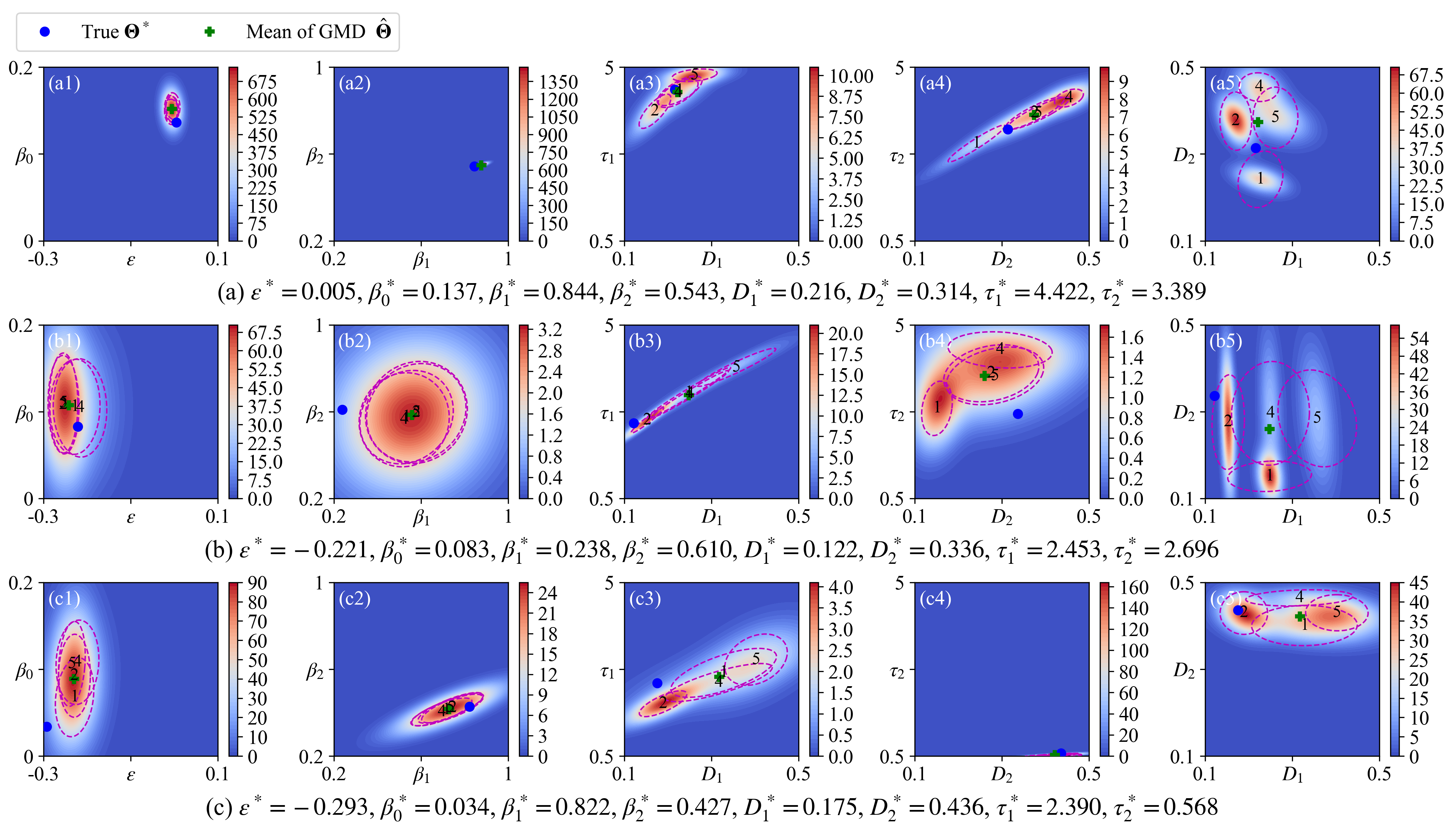}}
\caption{\label{fig:dvdp_distri}2-D marginal GMDs of several groups of parameter pairs provided by the PENN-GMD on the three test trajectories in Fig.~\ref{fig:dvdp_trajectories} of the DVDP in Eq.~(\ref{eq:sys_dvdp}). The 1-$\sigma$ ellipses of the Gaussian components with weights greater than 0.01 are drawn with labels.}
\end{figure}

To demonstrate that the GMD output of the PENN-GMD improves parameter identification and captures key couplings, we select the three trajectories shown in Fig.~\ref{fig:dvdp_trajectories} and examine several 2-D marginal distributions of the GMD estimates in Fig.~\ref{fig:dvdp_distri}. The first column of Fig.~\ref{fig:dvdp_distri} presents the marginal distributions of $\varepsilon$ and $\beta_0$. Only in Fig.~\ref{fig:dvdp_distri}(a1), the marginal mean is close to the true values and the marginal distribution is precise, indicating that the point estimate is accurate and the uncertainty is low. In contrast, the marginal distributions in Figs.~\ref{fig:dvdp_distri}(b1) and (c1) exhibit large variances, particularly for $\beta_0$. Although the true parameters deviate from the GMD means in these cases, they still fall within the high-probability regions, showing that the GMDs provide reasonable uncertainty estimates even when the point estimates are poor. Furthermore, the marginal distributions in Figs.~\ref{fig:dvdp_distri}(a1)-(a3) resemble vertical ellipses, suggesting that the estimates of $\varepsilon$ and $\beta_0$ are largely uncorrelated, which is consistent with the uniform error distribution observed in Fig.~\ref{fig:dvdp_estimates}(b1).

The second column of Fig.~\ref{fig:dvdp_distri} shows the marginal distributions of $\beta_1$ and $\beta_2$. Despite the dramatically different variances, the true values are all located in the respective high-probability regions, and these marginal distributions are positively correlated, consistent with our previous analysis. The third and fourth columns of Fig.~\ref{fig:dvdp_distri} exhibit the marginal distributions of $(D_1, \tau_1)$ and $(D_2, \tau_2)$, respectively. We can see that the two parameters in each noise source are positively correlated. The 1-D marginal distributions of $D_1$, $D_2$, $\tau_1$, and $\tau_2$ are wide, verifying the poor point estimates in Fig.~\ref{fig:dvdp_estimates}. Remarkably, their positively correlated joint distributions are very narrow and consistently contain the true parameters within their high-probability regions, indicating that the PENN-GMD can accurately identify the underlying relationship between the noise intensity and the correlation time for both the additive and multiplicative noises. This demonstrates that while the individual parameters may not be precisely recovered, their combination is well constrained. This underscores the need for a covariance-aware uncertainty representation. 

The last column of Fig.~\ref{fig:dvdp_distri} presents the marginal distributions of $D_1$ and $D_2$. In all three cases, the high-probability regions consist of several Gaussian components, and these distributions appear largely uncorrelated, indicating that the PENN-GMD can confidently identify both noise sources without confusing one for the other. 

\begin{figure}[!htb]
\center{\includegraphics[width=1\textwidth]
{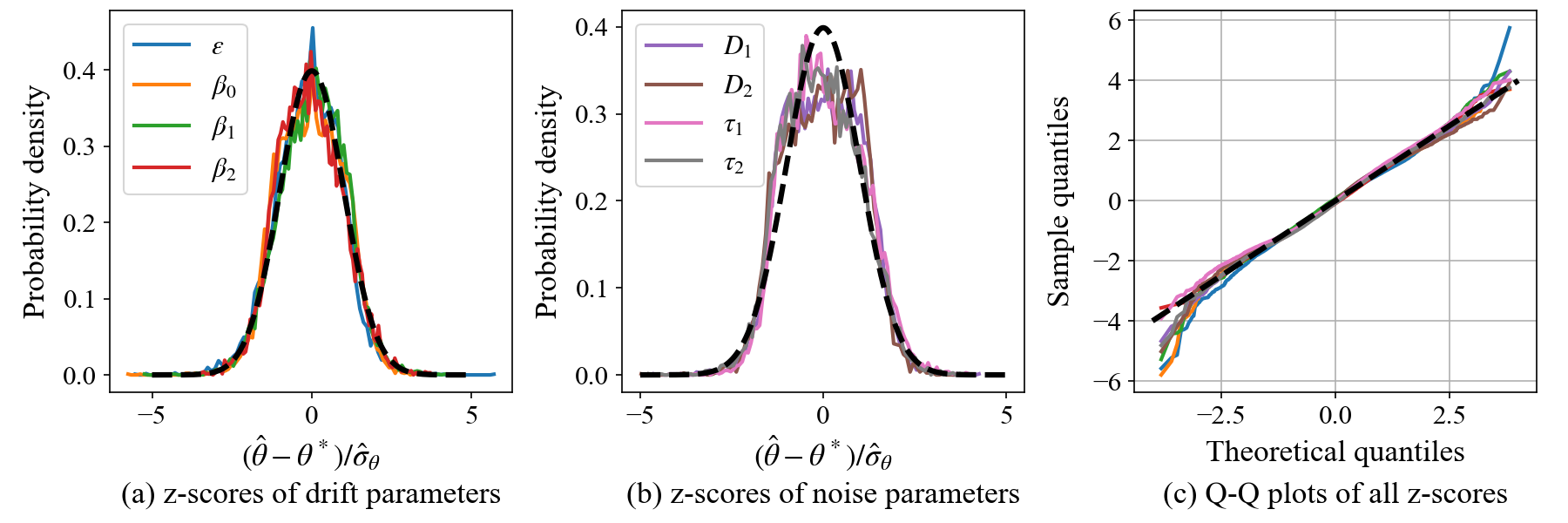}}
\caption{\label{fig:dvdp_zscore}Calibration of GMD marginal uncertainties on 10,000 test trajectories. (a)-(b) Histograms of $z$-scores for the four drift parameters (a) and four noise parameters (b). (c) Q-Q plots for all eight parameters. The black dashed lines denote perfect agreement with $\mathcal{N}(0,1)$.}
\end{figure}

Having established these qualitative patterns of the GMD distributions, we now turn to a quantitative validation of the underlying covariance structure. We process 10,000 trajectories and compute the total covariance matrices $\boldsymbol{\Sigma}_{\mathrm{GMD}}$ by Eq.~(\ref{eq:gmd_total_cov}). For each trajectory and each parameter $\theta$, we compute its standardized residual, i.e., the $z$-score $z=(\hat{\theta}-\theta^*)/\hat{\sigma}_{\theta}$~\cite{Rasmussen2023Uncertain}, where $\hat{\theta}$ is the point estimate and $\hat{\sigma}_{\theta}$ is the marginal estimation SD extracted from $\boldsymbol{\Sigma}_{\mathrm{GMD}}$. If the GMD variances are well-calibrated, these $z$-scores should follow the standard normal distribution $\mathcal{N}(0,1)$. Figs.~\ref{fig:dvdp_zscore}(a) and (b) show the histograms of $z$-scores aggregated over the four drift parameters and the four noise parameters, respectively, based on the 10,000 test trajectories. Both histograms closely match the theoretical $\mathcal{N}(0,1)$ density. The Q-Q plots in Fig.~\ref{fig:dvdp_zscore}(c) further confirm this alignment across all eight parameters, with the empirical quantiles falling on the 45$^\circ$ identity line within the central region $[-2.5,2.5]$. These results demonstrate that the diagonal entries of the GMD covariance matrix faithfully capture the point-estimation uncertainty of each individual parameter, establishing the statistical calibration of our uncertainty estimates.

\begin{figure}[!htb]
\center{\includegraphics[width=0.6\textwidth]
{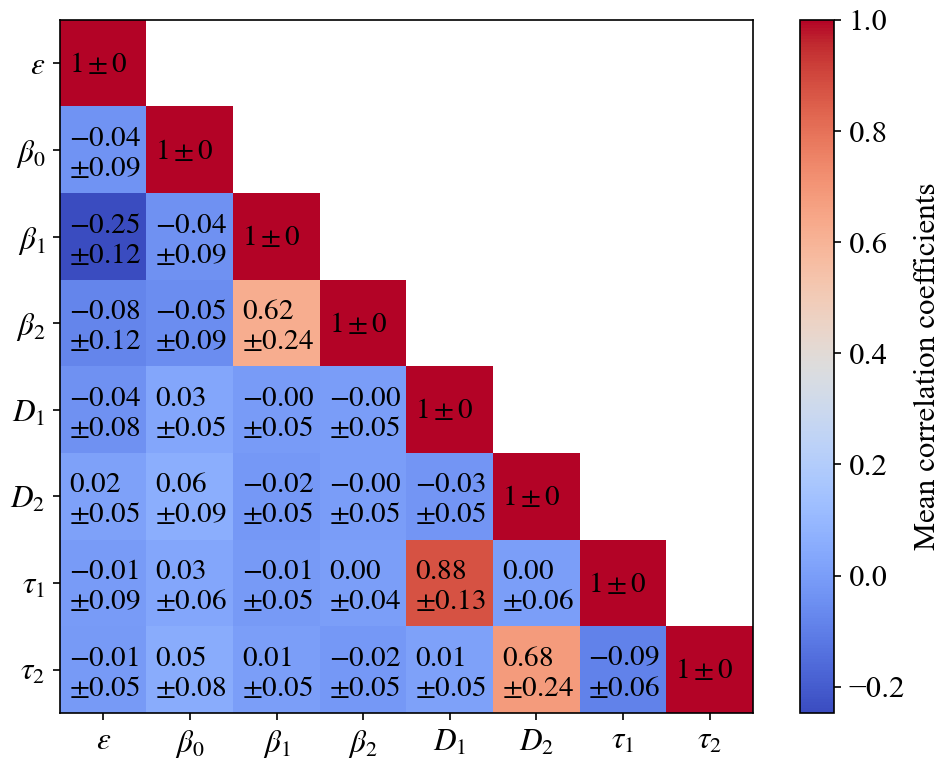}}
\caption{\label{fig:dvdp_corr}Means and SDs of the pairwise correlation coefficients among the eight DVDP parameters, estimated by PENN-GMD on 10,000 test trajectories with parameters uniformly sampled from $\mathcal{P}_\text{DVDP}$.}
\end{figure}

To examine the coupling structure captured by PENN-GMD more clearly, we report the pairwise correlation coefficients among the eight parameters in Fig.~\ref{fig:dvdp_corr}. Strong positive correlations are observed for the parameter pairs $(\beta_1,\beta_2)$, $(D_1,\tau_1)$, and $(D_2,\tau_2)$. A weak negative correlation between $\varepsilon$ and $\beta_1$ is also visible, which is consistent with the coupling term $\varepsilon+\beta_1 x^2$ in the system dynamics in Eq.~(\ref{eq:sys_dvdp}). In contrast, the estimates of $(D_1,D_2)$ and $(\tau_1,\tau_2)$ are essentially uncorrelated, indicating that the additive and multiplicative noise components can be identified without mutual interference. These observations confirm that PENN-GMD captures the coupling structure of the DVDP system parameters, revealing both the theoretically expected couplings rooted in the system dynamics and the independence between the two noise sources.

\subsection{Coupled FitzHugh-Nagumo neurons with L{\'e}vy noises}
\label{sec:cfhn}
We then investigate a couple of L{\'e}vy-driven FitzHugh-Nagumo neurons (CFHN)~\cite{Hoff2014Numerical}
\begin{equation}\label{eq:sys_cfhn}
\begin{split}
\dot{x}_1 &= c_1 \left( x_1 + y_1 - \frac{x_1^3}{3} \right) + \gamma_1 (x_1 - x_2), \\
\dot{y}_1 &= -\frac{1}{c_1}\left(x_1 - a_1 + b_1 y_1\right) + D_1^{1/\alpha_1}\dot{L}_{\alpha_1}(t), \\
\dot{x}_2 &= c_2 \left( x_2 + y_2 - \frac{x_2^3}{3} \right) + \gamma_2 (x_2 - x_1), \\
\dot{y}_2 &= -\frac{1}{c_2}\left(x_2 - a_2 + b_2 y_2\right) + D_2^{1/\alpha_2}\dot{L}_{\alpha_2}(t),\\
\end{split}
\end{equation}
with 12 parameters $\boldsymbol{\Theta}=[a_1, a_2, b_1, b_2, c_1, c_2, \gamma_1, \gamma_2, \alpha_1, \alpha_2, D_1, D_2]^\top$ in the parameter domain
\begin{equation}\label{eq:param_range_cfhn}
\begin{split}
\mathcal{P}_\text{CFHN}=&\{\boldsymbol{\Theta}|a_1, a_2\in[-3, 3];\;b_1,b_2\in[0, 3];\; c_1,c_2\in[0.1, 3];\; \gamma_1, \gamma_2 \in[-3, 3];\\
&\;\alpha_1,\alpha_2\in[1.2, 2];\;D_1,D_2\in[0.05, 0.5]\}.
\end{split}
\end{equation}
In this system, $x_i$ for $i=1,2$ is the membrane potential of the $i$-th neuron, and $y_i$ is the corresponding recovery variable. The parameters $a_i$, $b_i$, and $c_i$ are intrinsic parameters of each neuron. $a_i$ controls the excitability threshold, $b_i$ modulates the recovery time constant, and $c_i$ determines the time-scale separation between $x_i$ and $y_i$ variables. The coupling strengths $\gamma_1$ and $\gamma_2$ characterize the bidirectional synaptic interactions between the two neurons. Depending on their signs, positive couplings $\gamma_i > 0$ tend to amplify the difference between the two membrane potentials, suggesting inhibitory or anti-synchronizing interaction, while negative couplings $\gamma_i < 0$ tend to reduce the difference, leading to excitatory or synchronizing interaction. The present parameter domain $\mathcal{P}_\text{CFHN}$ allows both regimes. The L{\'e}vy noise terms $D_i^{1/\alpha_i}\dot{L}_{\alpha_i}(t)$ act independently on the recovery variables, where $\alpha_i$ and $D_i$ are the stability index and noise intensity of the symmetric $\alpha$-stable noise, respectively. As in Sec.~\ref{sec:oup}, these noise parameters are of particular interest, since they determine the non-Gaussian large jumps and intensity of the fluctuations driving each neuron.

We assume that only the membrane potentials $x_1(t)$ and $x_2(t)$ of the CFHN system are observable, while the recovery variables $y_1$ and $y_2$ remain hidden. This partial observability, together with the symmetry in the coupling structure, renders certain parameter combinations inherently challenging to identify, thereby providing a stringent test for the PENN-GMD framework. To systematically assess the impact of information availability, three observability scenarios are considered, corresponding to different choices of the measurement vector $\mathbf{z}$: (1) full observation ($\mathcal{O}_{x_1x_2}$) where $\mathbf{z} = [x_1, x_2]^\top$, (2) single-neuron observation ($\mathcal{O}_{x_1}$) where $\mathbf{z} = [x_1]$, and (3) mean observation ($\mathcal{O}_{\bar{x}}$) where $\mathbf{z} = [(x_1 + x_2)/2]$. The first scenario $\mathcal{O}_{x_1x_2}$ is the least challenging, as both membrane potentials are directly measured. In the second scenario $\mathcal{O}_{x_1}$, the parameters of the second neuron must be inferred indirectly from the first neuron's potential, making the problem considerably more difficult. Only the average signal is available in the third scenario $\mathcal{O}_{\bar{x}}$, which inherently mixes the contributions of the two neurons, leading to severe information mix-up and potential non-identifiability.

Beyond assessing the impact of observability, the GMD output of the PENN-GMD should also reveal intrinsic parameter couplings. For example, the time-scale $c_i$ and the noise intensity $D_i$ for $i=1,2$ exhibit a negative correlation. This correlation can be analytically illustrated for the Gaussian case $\alpha_i=2$. By replacing $x_i$ by a constant $\bar{x}_i$ in the $y_i$ dynamics, the recovery variable approximately follows
\begin{equation}
\dot{y}_i \approx \left(\frac{a_i-\bar{x}_i}{c_i}-\frac{b_i}{c_i} y_i\right) + \sqrt{2D_i}\dot{B}(t).
\end{equation}
This process has the same structural form as the SOUP in Eq.~(\ref{eq:soup}) up to a constant drift $(a_i-\bar{x}_i)/c_i$. The innovation variance of the exact AR(1) discretization, according to Eq.~(\ref{eq:ar1_var}), is $\sigma^2 = \frac{c_i}{b_i} D_i \left(1 - \text{e}^{-2\Delta t / (c_i/b_i)}\right)$. Thus, for a given observed fluctuation level, an increase in $c_i$ must be compensated by a decrease in $D_i$, giving rise to the negative correlation. This mechanistic trade-off is consistently observed in the joint marginal distributions in Fig.~\ref{fig:cfhn_distri}, confirming that the inferred negative correlation between $c_i$ and $D_i$ is physically grounded rather than spurious.

We generate 300,000 trajectories of Eq.~(\ref{eq:sys_cfhn}) using the Euler-Maruyama method with integration step $\delta t = 0.01$ and sampling time $\Delta t=0.2$. Each trajectory has $L=1500$ samples, with a random burn-in of $t_0 = k\Delta t$ where $k$ is uniformly sampled in the interval $[500,1500]$ to diversify the initial conditions seen by the network and improve generalization. We deliberately use long trajectories in the training and test stages to ensure that the trajectories are informative enough, such that uncertainties are mainly structural. Three PENN-GMDs are trained from the partially observed 2-D, 1-D, and 1-D trajectories with the three observability scenarios $\mathcal{O}_{x_1x_2}$, $\mathcal{O}_{x_1}$, and $\mathcal{O}_{\bar{x}}$, respectively, using batch size $N_\text{batch}=1200$ for $N_\text{epoch}=3000$ epochs, where the validation loss converges.

\begin{figure}[!htb]
\center{\includegraphics[width=1\textwidth]
{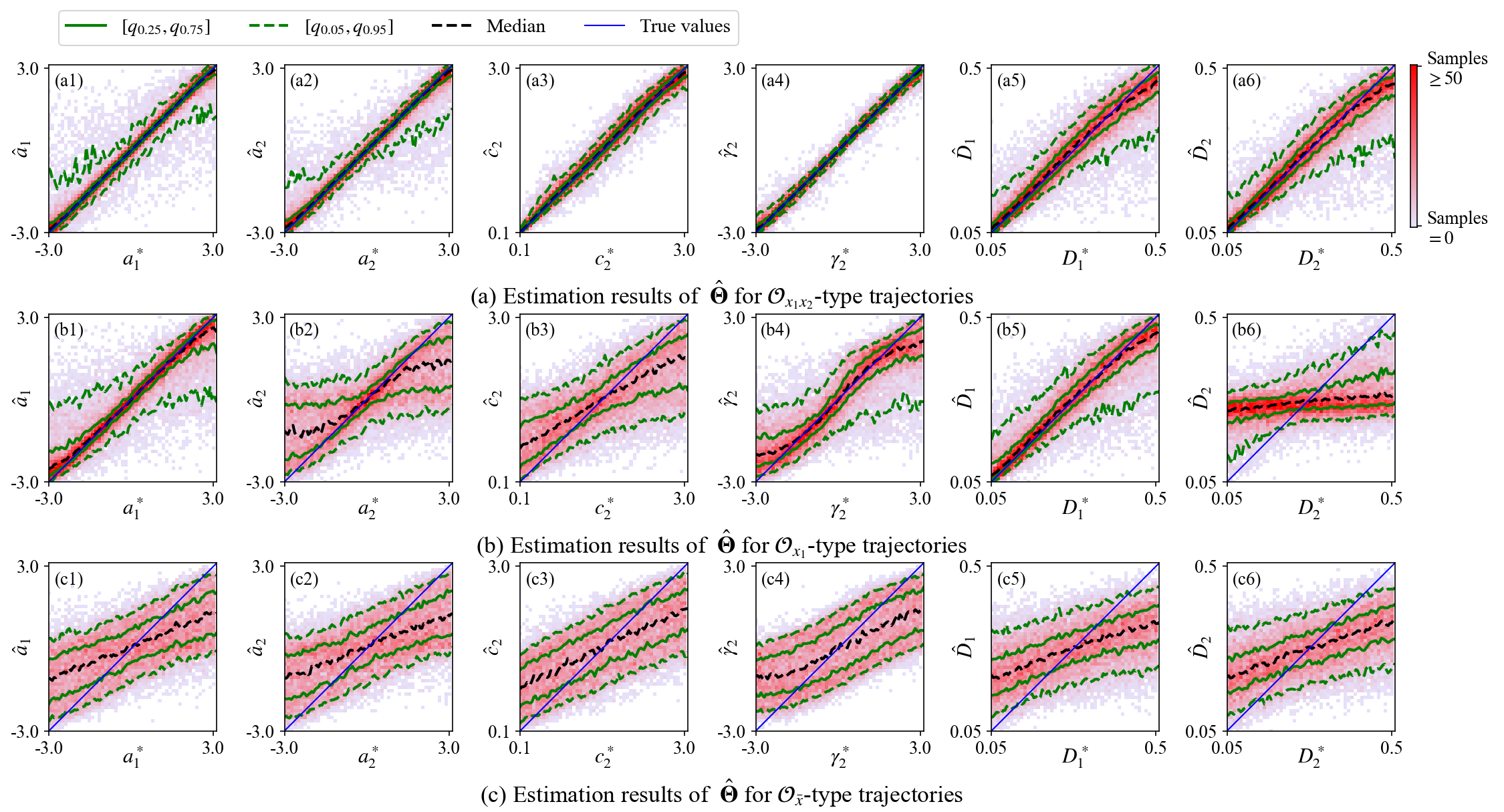}}
\caption{\label{fig:cfhn_estimates}The histograms and quantiles of point estimates versus the true values of selected parameters by the PENN-GMD on 20,000 test trajectories of the CFHN system in Eq.~(\ref{eq:sys_cfhn})}
\end{figure}

Fig.~\ref{fig:cfhn_estimates}(a) shows the point estimates of six parameters from 20,000 test trajectories under $\mathcal{O}_{x_1x_2}$. Overall, all six parameters are identifiable, with their 25\%-75\% quantiles lying near the diagonals. The estimates of $D_1$ and $D_2$ exhibit relatively large variances, reflecting the inherent difficulty in estimating noise intensities. Due to the symmetry between the two neurons in both the equations and the parameter ranges, the estimates of $a_1$ and $D_1$ in Figs.~\ref{fig:cfhn_estimates}(a1) and (a5) are identical to those of $a_2$ and $D_2$ in Figs.~\ref{fig:cfhn_estimates}(a2) and (a6), respectively. Fig.~\ref{fig:cfhn_estimates}(b) plots the point estimates under $\mathcal{O}_{x_1}$ observability. Compared with the counterparts in Figs.~\ref{fig:cfhn_estimates}(a), the estimation accuracies of $a_1$, $D_1$ for the first neuron degrade only slightly without the measurements of $x_2$. In contrast, all parameters of the second neuron become significantly harder to identify, with $D_2$ being nearly non-identifiable as the point estimates $\hat{D}_2$ significantly shrink toward the center of the parameter range $[0.05, 0.5]$ in $\mathcal{P}_\text{CFHN}$. This asymmetry arises because information about the second neuron reaches the observed signal $x_1$ only through the coupling term $\gamma_1(x_1-x_2)$.

The point estimates from the mean observation scenario $\mathcal{O}_{\bar{x}}$ are shown in Fig.~\ref{fig:cfhn_estimates}(c), where all estimates are poor due to a structural limitation for symmetric systems. When only the average potential $(x_1+x_2)/2$ is observed, the two neurons become exchangeable. For a trajectory with true parameters $(\theta_1^*, \theta_2^*)$, where $\theta$ denotes any of the six parameters associated with each individual neuron, a capable estimator such as the PENN-GMD assigns comparable weights to the two symmetric modes $(\theta_1^*, \theta_2^*)$ and $(\theta_2^*, \theta_1^*)$. The point estimate, defined as the weighted mean of these modes, falls near the diagonal $((\theta_1^*+\theta_2^*)/2, (\theta_1^*+\theta_2^*)/2)$. In other words, for each neuron-specific parameter $\theta_k$ for $k=1,2$, the point estimate $\hat{\theta}_k$ is approximately $(\theta_1^*+\theta_2^*)/2$, with the deviation from $\theta_k^*$ being roughly proportional to $\theta_1^*-\theta_2^*$. Consequently, when the true parameters are uniformly sampled from the square domain $\mathcal{P}_{\mathrm{CFHN}}$, the relationship between $\theta_k^*$ and $\hat{\theta}_k$ collapses the square onto a parallelogram centered at the origin in each $(\theta_k^*, \hat{\theta}_k)$ plane. The diagonal with low bias corresponds to $\theta_1^* = \theta_2^*$, where the estimate matches the true value. The other diagonal with maximal bias corresponds to the two corners $(\theta_1^{\min}, \theta_2^{\max})$ and $(\theta_1^{\max}, \theta_2^{\min})$, where the estimate is pinned at the center $(\theta_1^{\min}+\theta_1^{\max})/2$ and the error reaches $(\theta_1^{\max}-\theta_1^{\min})/2$ in magnitude. This parallelogram structure is precisely observed in the distributions of Fig.~\ref{fig:cfhn_estimates}(c), confirming that the failure of the point estimate under $\mathcal{O}_{\bar{x}}$ is not random but systematically governed by the exchangeability symmetry. This limitation, inherent to any single-point estimator, underscores the necessity of the multi-modal GMD representation.

\begin{figure}[!htb]
\center{\includegraphics[width=1\textwidth]
{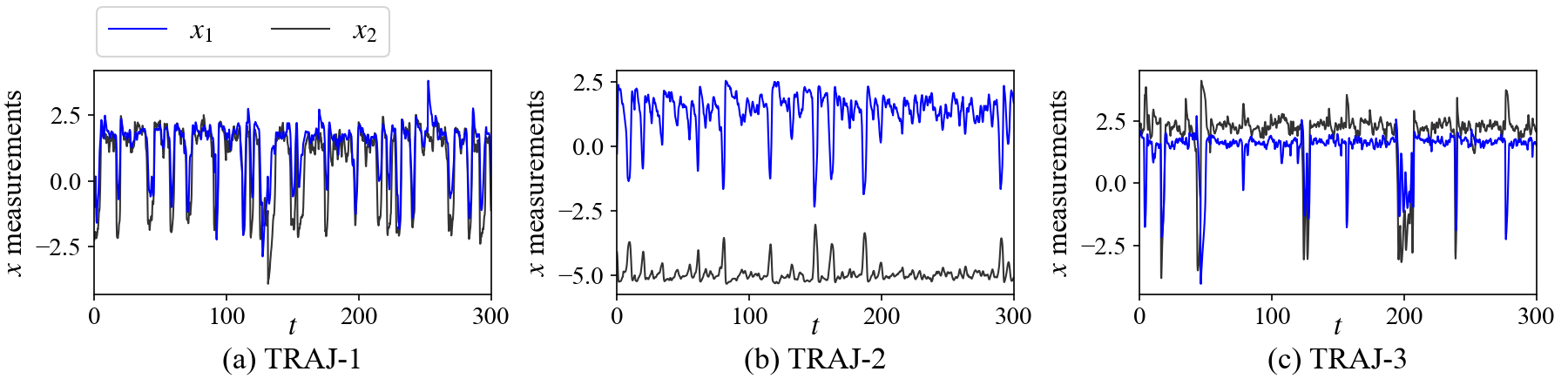}}
\caption{\label{fig:cfhn_trajectories}The $x_1$ and $x_2$ measurements of three trajectories of the CFHN system in Eq.~(\ref{eq:sys_cfhn}). The system parameters and the GMD estimations by the PENN-GMD are detailed in Fig.~\ref{fig:cfhn_distri}.}
\end{figure}

\begin{figure}[!htb]
\center{\includegraphics[width=1\textwidth]
{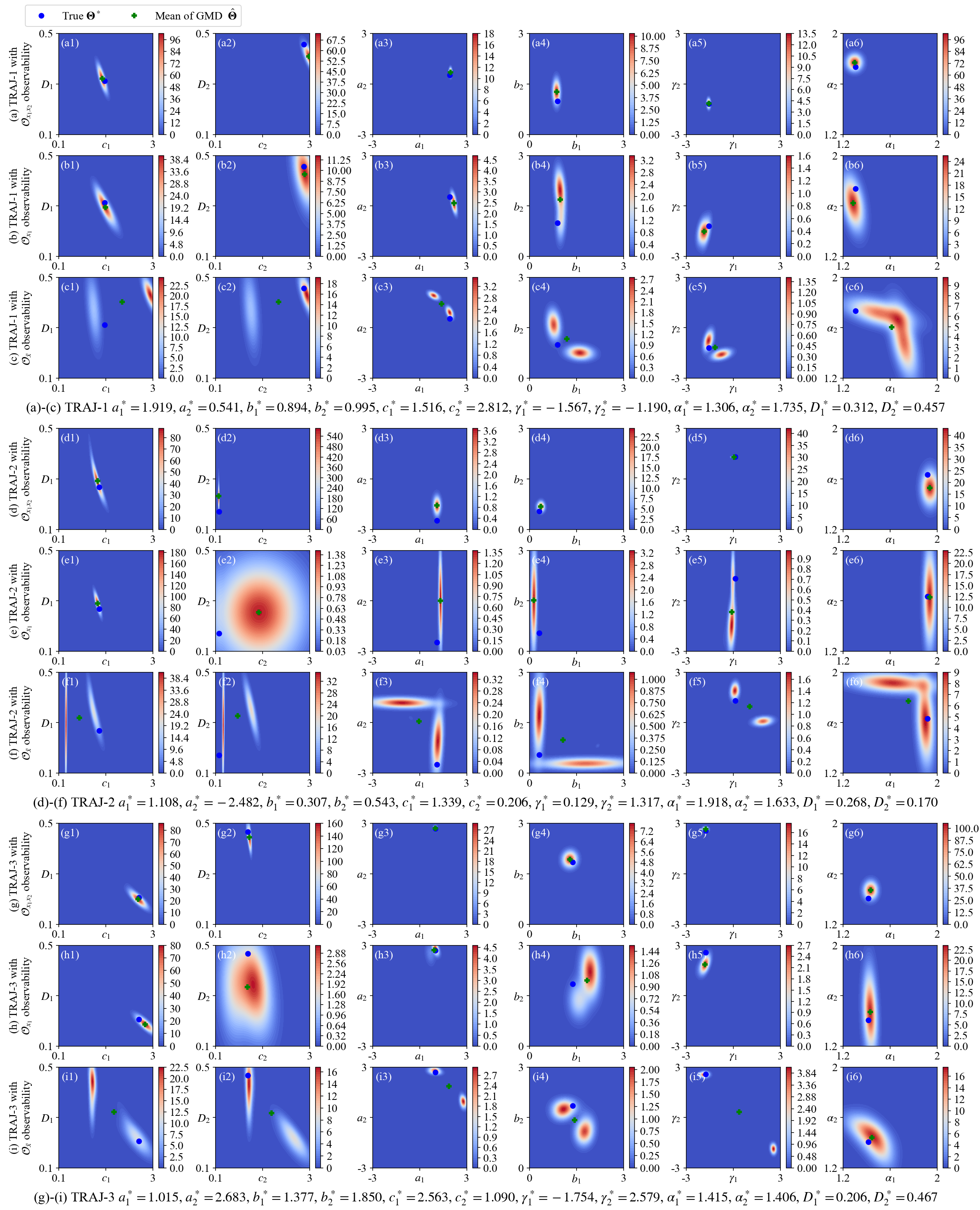}}
\caption{\label{fig:cfhn_distri}2-D marginal GMDs obtained by the PENN-GMDs for the three trajectories in Fig.~\ref{fig:cfhn_trajectories} under different observability types.}
\end{figure}

We then evaluate the trained PENN-GMDs on the three representative trajectories displayed in Fig.~\ref{fig:cfhn_trajectories}. The trajectory TRAJ-1 in Fig.~\ref{fig:cfhn_trajectories}(a) corresponds to negative coupling strengths, under which $x_1$ and $x_2$ become synchronized. TRAJ-2 in Fig.~\ref{fig:cfhn_trajectories}(b) has positive couplings, which drive $x_1$ and $x_2$ to evolve in opposite directions. TRAJ-3 Fig.~\ref{fig:cfhn_trajectories}(c) features a mixed coupling configuration, i.e., $\gamma_1<0$, $\gamma_2>0$, resulting in a chase-run dynamics between the two variables.

Nine groups of 2-D marginal GMDs are illustrated in Fig.~\ref{fig:cfhn_distri}, corresponding to the three PENN-GMDs on the three trajectories with different observability scenarios. Fig.~\ref{fig:cfhn_distri}(a) shows the marginal GMDs of TRAJ-1 under $\mathcal{O}_{x_1x_2}$ observability. When both membrane potentials can be observed, the point estimates are accurate and the uncertainties of all distributions are low. We can also see that $c_i$ and $D_i$ are negatively correlated in Figs.~\ref{fig:cfhn_distri}(a1) and (a2). The pairs $(a_1, a_2)$ and $(b_1, b_2)$, $(\gamma_1, \gamma_2)$, and $(D_1, D_2)$ in Figs.~\ref{fig:cfhn_distri}(a3)-(a6) are uncorrelated. These patterns confirm that the PENN-GMD not only gives accurate and precise estimates, but also correctly recovers the couplings among the parameters of each FHN as well as the absence of couplings between parameters of different FHNs.

Fig.~\ref{fig:cfhn_distri}(b) details the marginal distributions of TRAJ-1 under $\mathcal{O}_{x_1}$ observability. Compared with the counterparts in Figs.~\ref{fig:cfhn_distri}(a), the GMDs in Figs.~\ref{fig:cfhn_distri}(b) have larger variances, due to the information in $x_2$ measurements is missing. The increases of uncertainties are significantly larger for the parameters of the second FHN, i.e., Fig.~\ref{fig:cfhn_distri}(b2) and the vertical axes of Figs.~\ref{fig:cfhn_distri}(b3)-(b6), while the variances of the parameters of the first FHN only increase slightly, i.e., Fig.~\ref{fig:cfhn_distri}(b1) and the horizontal axes of Figs.~\ref{fig:cfhn_distri}(b3)-(b6).

Fig.~\ref{fig:cfhn_distri}(c) demonstrates the marginal distributions of TRAJ-1 under $\mathcal{O}_{\bar{x}}$ observability. The majority of the marginal distributions are bimodal and the two modes are symmetric in that exchanging the parameter indices of the two FHNs results in the same GMD. This excellent result indicates that the PENN-GMD accurately identifies all the 12 parameters, since the true values are always located in one of the two high-probability regions, except that it can not distinguish the order of the two FHNs. Therefore, it produces a likelihood with both possibilities.

Similar patterns can be seen in Figs.~\ref{fig:cfhn_distri}(d)-(f) of TRAJ-2 and (g)-(i) of TRAJ-3, where the three PENN-GMDs confidently identify all the 12 parameters with modes concentrated near the true values and densities showing reasonable uncertainties, couplings, and symmetries. One can also see why the point estimate fails in the $\mathcal{O}_{\bar{x}}$ scenario where the PENN-GMD succeeds. In Fig.~\ref{fig:cfhn_distri}(f6), the accurate likelihood provided by the PENN-GMD includes horizontal and vertical modes, corresponding to the belief that one stability index is precise and the other is uncertain. The point estimate $\hat{\boldsymbol{\Theta}}$, which is the mean of the two modes, has very low probability density. This is the reason for the high errors in Fig.~\ref{fig:cfhn_estimates}(c). In Fig.~\ref{fig:cfhn_distri}(i6), the distribution of the stability indices $(\alpha_1, \alpha_2)$ of TRAJ-3 under $\mathcal{O}_{\bar{x}}$ observability is unimodal instead of bimodal, since the true values $\alpha_1=1.414$ and $\alpha_2=1.406$ are close. In such cases the two modes are fused into one and the point estimate happens to be accurate.

\subsection{An airfoil system with Gaussian and L{\'e}vy-driven colored noise}
\label{sec:airfoil}

We finally apply the proposed method to an aeroelastic pitch-plunge airfoil~\cite{Liu2021Fixed,Feng2025Fusing}, a canonical model in nonlinear aeroelasticity that captures the essential dynamics of flutter and limit-cycle oscillations (LCOs) in flexible aircraft system. Accurate identification of its structural and aerodynamic parameters from limited measurements is critical for stability assessment and safe flight envelope prediction, yet remains challenging due to strong nonlinearities, partial observability, and uncertain turbulent excitations. The airfoil system is described by the coupled bending-torsion non-dimensional equations:
\begin{equation}\label{eq:airfoil}
\begin{split}
\ddot{\xi} + x_\varphi \ddot{\varphi} +& 2\zeta_\xi(\bar{\omega}/U)\dot{\xi} + (\bar{\omega}/U)^2G(\xi)=-C_L(t,a_h)/(\pi\mu),\\
(x_\varphi/r_\varphi^2)\ddot{\xi}+\ddot{\varphi}+&2\zeta_\varphi(1/U)\dot{\varphi}+(1/U)^2M(\varphi)=2C_M(t,a_h)/(\pi\mu r_\varphi^2).
\end{split}
\end{equation}
Here, $\xi$ is the non-dimensional plunge displacement and $\varphi$ is the pitch angle about the elastic axis. $G(\xi)=\xi + \gamma_3\xi^3$ and $M(\varphi)=\varphi + \beta_3\varphi^3$ are the nonlinear cubic stiffnesses of plunge and pitch, respectively. $C_L$ and $C_M$ are the lift and torque moment, respectively. This deterministic system is governed by the non-dimensional flow velocity $U$, nonlinear stiffness parameters $\beta_3$ and $\gamma_3$, damping parameters $\zeta_\xi$ and $\zeta_\varphi$, the radius of gyration about the elastic axis $r_\varphi$, the distances ratios $a_h$ of the mid-chord and $x_\varphi$ of the center mass from the elastic axis, and the airfoil-air mass ratio $\mu$. 

We further assume that the flow velocity $U(t)=U_0+\eta(t)$ is a base flow velocity $U_0$ disturbed by a Gaussian and L{\'e}vy-driven colored noise $\eta(t)$, which captures both small fluctuations as well as abrupt gusts. Similar to Eq.~(\ref{eq:ou_fgn_levy}), $\eta(t)$ is a mixed noise-driven OUP
\begin{equation}\label{eq:ou_gaussian_levy}
\dot{\eta}(t) = -\frac{1}{\tau} \eta(t) + \sqrt{\frac{2D_{\mathrm{Gauss}}}{\tau}}\dot{B}(t) + \left(\frac{D_{\mathrm{L\acute{e}vy}}}{\tau}\right)^{1/\alpha}\dot{L}_\alpha(t),
\end{equation}
with correlation time $\tau$, intensities $D_{\mathrm{Gauss}}$ of the Gaussian noise and $D_{\mathrm{L\acute{e}vy}}$ of the L{\'e}vy noise, and the stability index $\alpha$ of the L{\'e}vy noise. The airfoil system is governed by 11 parameters $\boldsymbol{\Theta}=[U_0, \beta_3, \gamma_3, x_\varphi, r_\varphi, a_h, \mu, \tau, D_\mathrm{Gauss}, D_{\mathrm{L\acute{e}vy}}, \alpha]^\top$ with the parameter domain
\begin{equation}
\begin{split}
\mathcal{P}_\text{Airfoil}=\{\boldsymbol{\Theta}|& U_0\in[5,16]; \beta_3\in[40,120]; \gamma_3\in[5,15]; x_\varphi\in[0.2,0.3]; r_\varphi\in[0.4,0.6];\\ a_h\in[-0.6, -0.4]; & \mu\in[80,120]; \tau\in[0.5, 5]; D_\mathrm{Gauss}\in[0, 0.1]; D_{\mathrm{L\acute{e}vy}}\in[0, 0.1]; \alpha\in[1.2, 2]\}.
\end{split}
\end{equation}
The ranges of the seven airfoil parameters are classic choices~\cite{Feng2025Fusing,Feng2026Data}. To make the scenario more practical, we treat the damping parameters $\zeta_\xi\in[0, 0.05]$, $\zeta_\varphi\in[0,0.05]$ as unobservable trajectory-dependent parameters in the simulation. These parameters are outside our estimation scope but influence the measurement data. Therefore, we only identify a subset of all governing parameters.

To simulate the airfoil system, Eq.~(\ref{eq:airfoil}) is transformed into a group of first-order ordinary differential equations with the 6-D state $\mathbf{x}(t)=[\varphi(t),\dot{\varphi}(t), \xi(t),\dot{\xi}(t),\omega_1(t),\omega_2(t)]^\top$, where $\omega_1$ and $\omega_2$ are auxiliary variables to eliminating the integro-differential terms~\cite{Feng2025Fusing}. We assume that only the 2-D pitch and plunge measurements are observable, i.e., $\mathbf{z}=[\varphi, \xi]^\top$, leading to an identification problem with almost a dozen parameters, strong nonlinearity, non-Gaussian mixed-noise, and partial observability.

A PENN-GMD is trained on 300,000 trajectories using the 4th-order Runge-Kutta (RK4) method while the OUP is simulated by Euler-Maruyama method. The integration step is $\delta t=0.1$ and sampling time is $\Delta t=0.5$. The samples within $t_0=1000\Delta t$ are omitted to remove transients. Each trajectory includes $L=1000$ measurements of $\mathbf{z}=[\varphi, \xi]^\top$ and the PENN-GMD is trained for 5000 epochs.

\begin{figure}[!htb]
\center{\includegraphics[width=1\textwidth]
{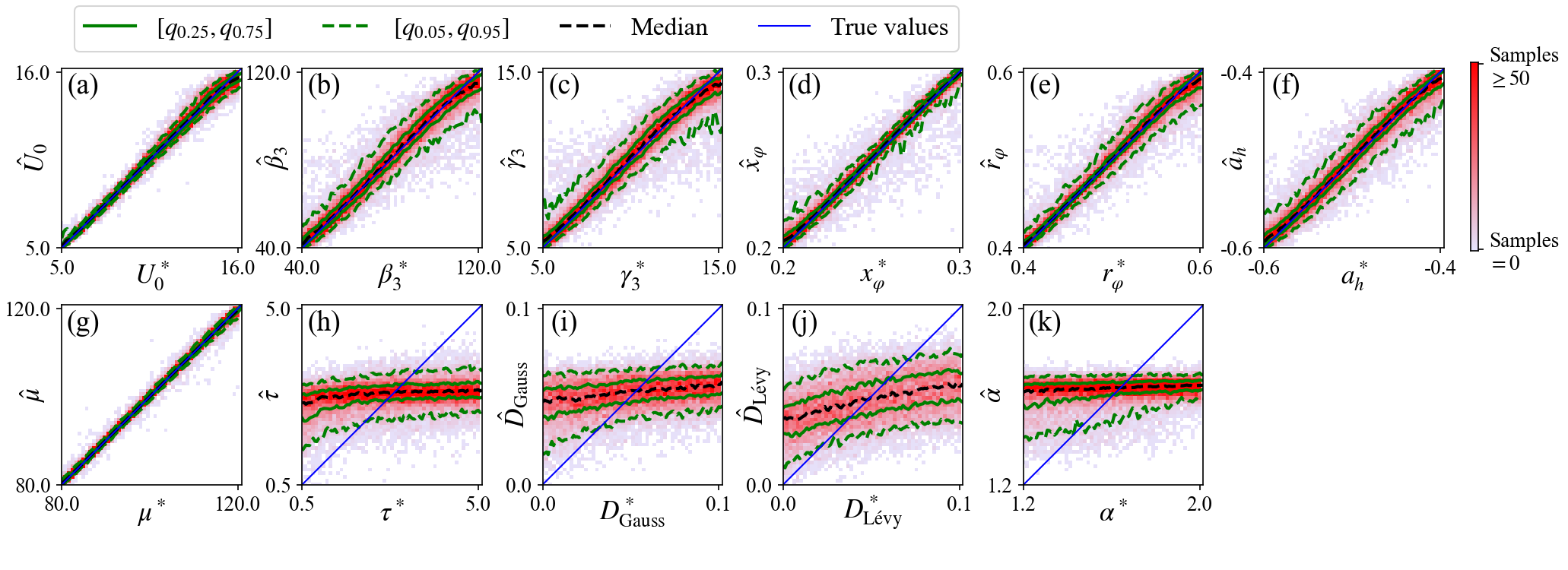}}
\caption{\label{fig:airfoil_estimates}Point estimates by PENN-GMD on 20,000 test trajectories of the stochastic airfoil system in Eqs.~(\ref{eq:airfoil}) and (\ref{eq:ou_gaussian_levy}).}
\end{figure}

Fig.~\ref{fig:airfoil_estimates} shows the point estimates of PENN-GMD. The first seven airfoil parameters can be effectively identified. The base flow velocity $U_0$ and the airfoil-air mass ratio $\mu$ are estimated the most accurate while the variances of the stiffness parameters $\beta_3$, $\gamma_3$ and structural parameters $x_\varphi$, $r_\varphi$, $a_h$ are larger. In contrast, the four noise parameters $\tau$, $D_\text{Gauss}$, $D_{\mathrm{L\acute{e}vy}}$, and $\alpha$ are practically non-identifiable from the available measurements. The PENN-GMD only results values near the centers of their sampling ranges. This result is reasonable as the colored noise is weak and only affects the flow velocity, which in turn influences the pitch and plunge in parts of Eq.~(\ref{eq:airfoil}). The great uncertainty indicates that the point estimates of the noise parameters are extremely unreliable.

\begin{figure}[!htb]
\center{\includegraphics[width=0.7\textwidth]
{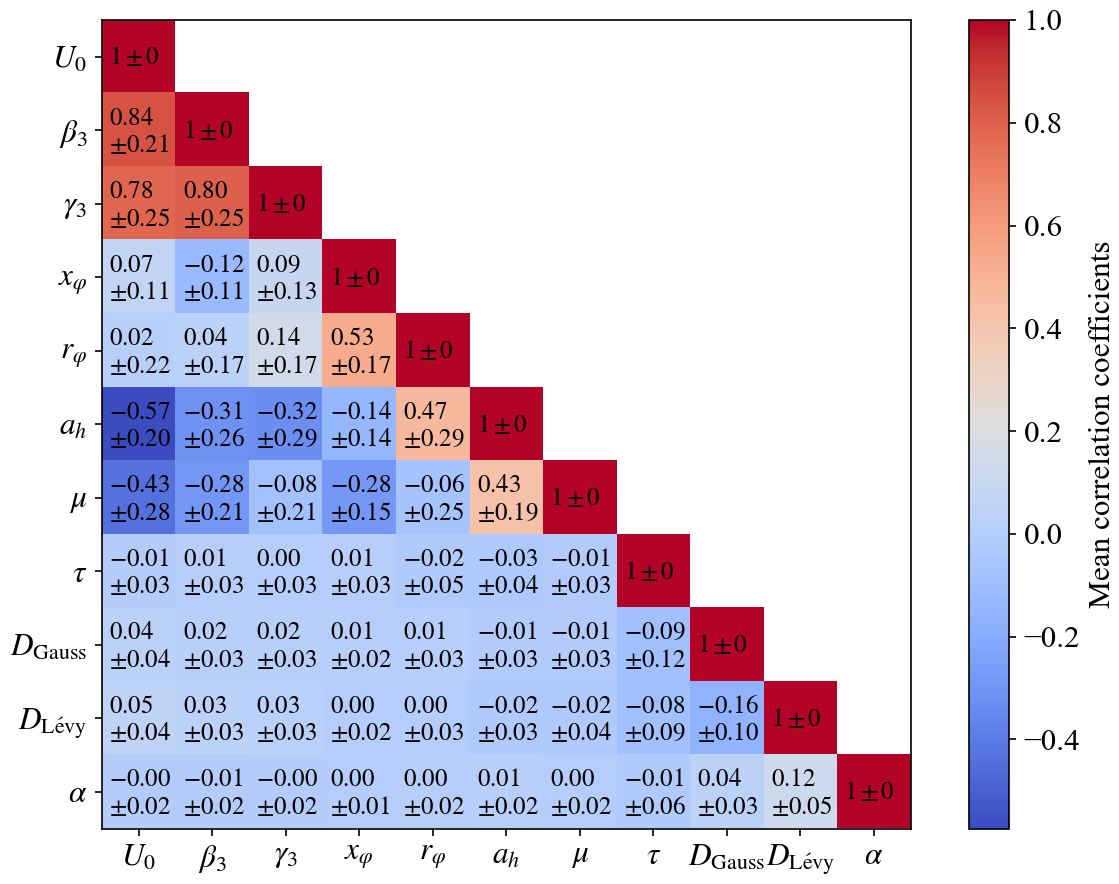}}
\caption{\label{fig:airfoil_correlations}Means and SDs of the pairwise correlation coefficients among the eleven parameters of the stochastic airfoil, estimated by PENN-GMD on 20,000 test trajectories with parameters uniformly sampled from $\mathcal{P}_\text{Airfoil}$.}
\end{figure}

By using PENN-GMD, the couplings among parameters are partially revealed via the total covariance matrix $\boldsymbol{\Sigma}_{\mathrm{GMD}}$ computed from 20,000 test trajectories, as shown in Fig.~\ref{fig:airfoil_correlations}. The first seven airfoil parameters are largely uncorrelated with the four noise parameters. The base flow velocity $U_0$ is positively correlated with the stiffness parameters $\beta_3$ and $\gamma_3$, reflecting that higher flow velocities generate larger aerodynamic loads, which must be balanced by stronger structural nonlinearities to maintain LCOs. The positive correlation between $x_\varphi$ and $r_\varphi$ arises partially from the coupling term $x_\varphi/r_\varphi^2$ in Eq.~(\ref{eq:airfoil}), correctly indicating the geometric dependency of the airfoil structure. The negative mean correlation $-0.57$ between $U_0$ and $a_h$, and the positive mean correlation $0.47$ between $a_h$ and $\mu$, are also physically consistent. A more aft elastic axis or more negative $a_h$ enhances pitch excitation efficiency, requiring a smaller base flow velocity $U_0$, while a larger mass ratio $\mu$ weakens aerodynamic forcing, necessitating a more aft elastic axis to compensate. Together, these correlations reveal that $a_h$ serves as a pivotal parameter bridging aerodynamic excitation and structural response, and their joint structure is faithfully captured by $\boldsymbol{\Sigma}_{\mathrm{GMD}}$.

Though the point estimate fails to identify the noise parameters, the GMD captures their weak correlations. Fig.~\ref{fig:airfoil_correlations} shows that the noise intensities $D_\mathrm{Gauss}$ and $D_{\mathrm{L\acute{e}vy}}$ are negatively correlated (mean $-0.16$), indicating their joint contribution to fluctuations. The stability index $\alpha$ and intensity $D_{\mathrm{L\acute{e}vy}}$ are positively correlated (mean $0.12$), since a larger $\alpha$ reduces jump amplitudes and requires a larger intensity to maintain the same fluctuation level.

\begin{figure}[!htb]
\center{\includegraphics[width=1\textwidth]
{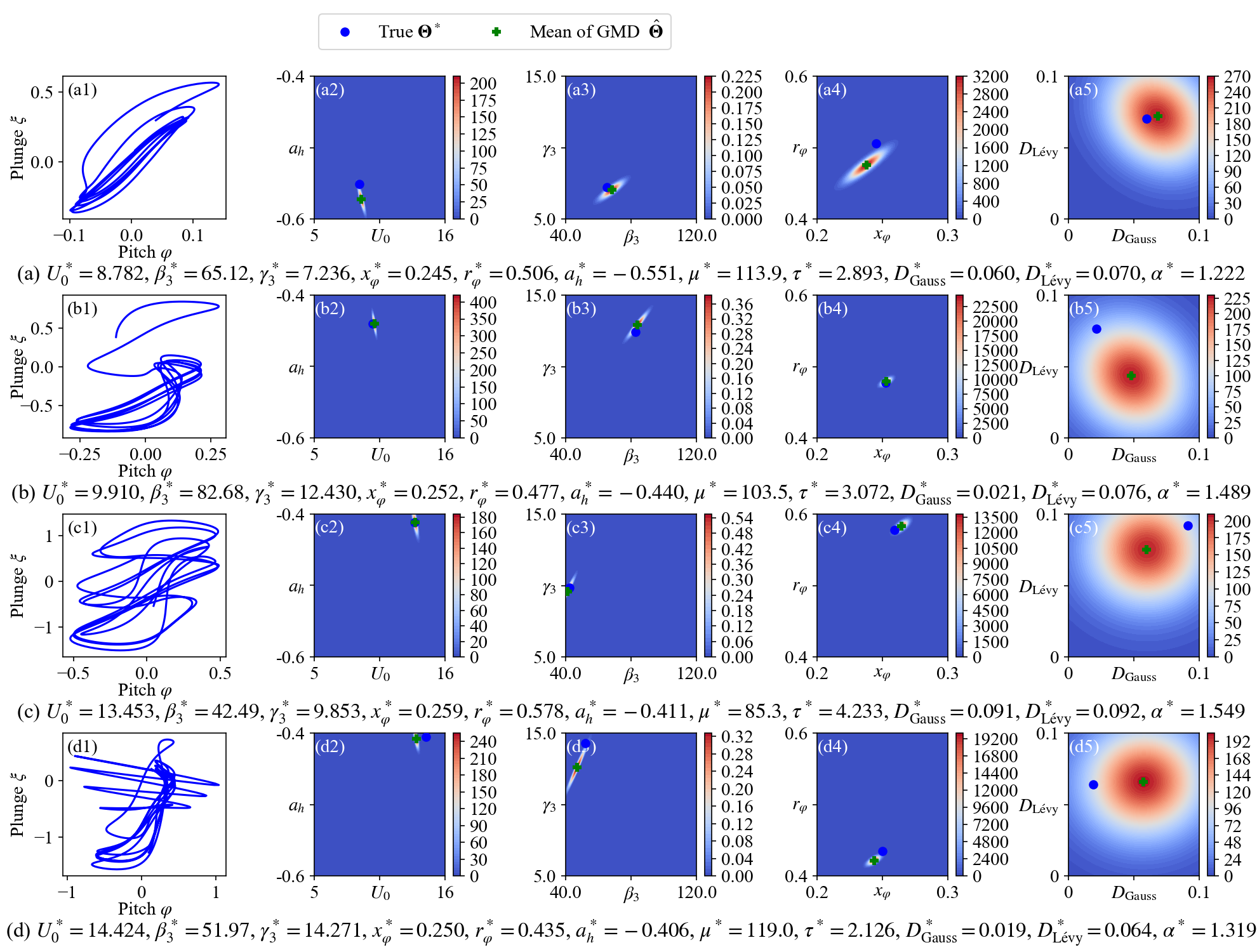}}
\caption{\label{fig:airfoil_gmds}2-D marginal GMDs obtained by PENN-GMD for four trajectories of the stochastic airfoil system.}
\end{figure}

Three test trajectories of the pitch and plunge measurements are plotted in the first column of Fig.~\ref{fig:airfoil_gmds} and several 2-D marginal GMDs are shown. PENN-GMD correctly recognizes the strongly negative correlations of $(U_0,a_h)$, strongly positive correlations of $(\beta_3,\gamma_3)$ and $(x_\varphi, r_\varphi)$, and the weakly negative correlation of $(D_\mathrm{Gauss},D_{\mathrm{L\acute{e}vy}})$. It can be seen that the estimates of the system parameters in the middle three columns are very accurate and precise and the estimates of the noise intensities in the last column have huge variances, indicating great uncertainties. 

This airfoil case demonstrates the engineering value of the PENN-GMD framework in three aspects. First, despite strong nonlinearity, partial observability, and an unmodeled damping perturbation, the seven physically meaningful airfoil parameters are reliably identified, with their uncertainties faithfully quantified by the GMD covariance. Second, the total covariance matrix automatically uncovers the underlying physical compensation mechanisms among parameters, such as the trade-off between flow velocity and cubic stiffness, offering interpretable insights that point estimates cannot provide. Third, when the data contain insufficient information to resolve the noise parameters, the GMD does not produce overconfident point estimates but instead reports broad distributions and weak correlations, providing honest uncertainty quantification that is critical for risk-informed engineering decisions. These results confirm that PENN-GMD is not merely a parameter estimator, but a diagnostic tool that reveals the identifiability structure of complex aeroelastic systems from limited, noise-corrupted observations.

\section{Discussion}
\label{sec:discussion}
\begin{figure}[!htb]
\center{\includegraphics[width=1\textwidth]
{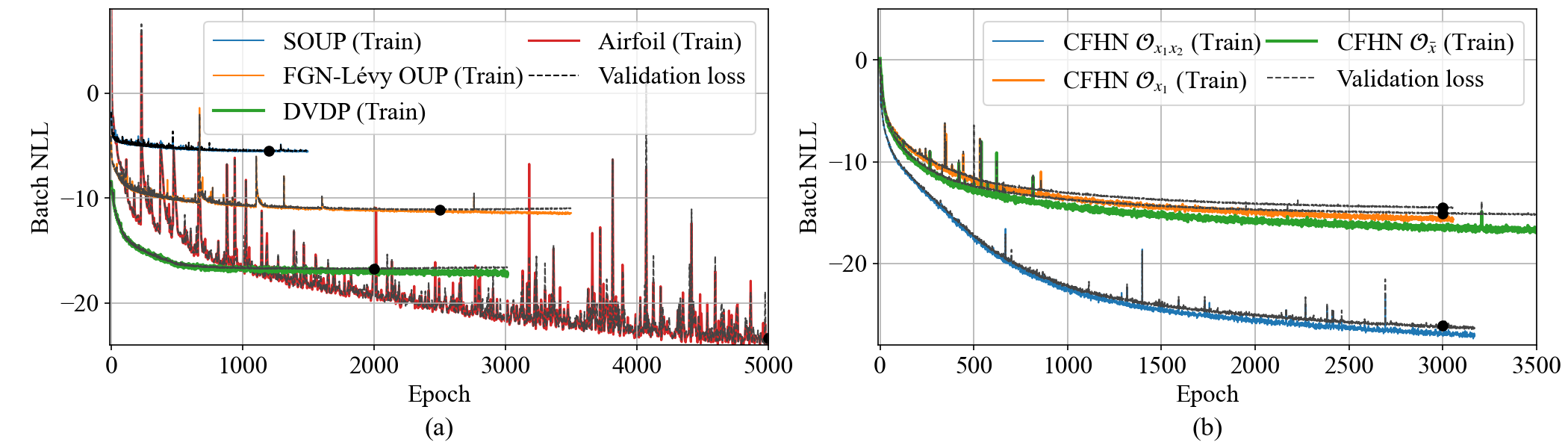}}
\caption{\label{fig:loss_curves}The loss curves of the PENN-GMDs in Sec.~\ref{sec:numer_ex}. Each PENN-GMD includes a training (train) curve calculated the last batch of each training epoch and a validation (val) curve averaged on an independent dataset with 20,000 trajectories. The black dots indicate the model for testing.}
\end{figure}

The training of PENN-GMD is simple as the NLL loss in Eq.~(\ref{eq:batch_loss}) has no hyperparameters. Fig.~\ref{fig:loss_curves} shows the loss curves of the PENN-GMDs in Sec.~\ref{sec:numer_ex}. The loss curves of the first three systems in Sec.~\ref{sec:soup} to Sec.~\ref{sec:dvdp} and the airfoil in Sec.~\ref{sec:airfoil} are detailed in Fig.~\ref{fig:loss_curves}(a), and those of the CFHN systems with different observability scenarios are shown in Fig.~\ref{fig:loss_curves}(b). Each example includes a training curve on 300,000 or 500,000 trajectories and a validation curve on 20,000 trajectories. In all the seven PENN-GMDs, the training curves are close to the corresponding validation curves, indicating that the training data are sufficient to avoid severe overfitting. Though there are occasional fluctuations due to the stochastic training data, the large deviations always quickly evolve to the main curves. These observations confirm that the training process of the PENN-GMD is stable. Tab.~\ref{tab:config} details the computational cost of PENN-GMDs. Though the training takes several hours, the inference speed is very high. As the map from trajectories to GMD parameters can be evaluated in parallel on GPU, more than 20,000 trajectories can be identified within a second. Therefore, once the training is completed, the PENN-GMD can identify trajectories in real-time.

It is worth noting that data generation is also highly efficient with GPU acceleration. The RK4 integrator and the CMS algorithm for generating L{\'e}vy noises are implemented in PyTorch~\cite{pytorch2019}, enabling parallel trajectory generation. For instance, generating 100,000 trajectories of the stochastic airfoil system with trajectory-dependent parameters consumes approximately 3 GiB of GPU memory and takes about 48 seconds. Of this, 37 seconds are spent on warm-up and transient removal, while the actual recording of the trajectories takes only 8 seconds. This efficiency provides a virtually unlimited data supply for training.

\begin{table}[htbp]
\footnotesize
\centering
\caption{Computational costs for the five numerical examples in Sec.~\ref{sec:numer_ex}. Training and inference times are estimated on a single NVIDIA RTX 4090 GPU.}
\label{tab:config}
\begin{tabular}{l c c c c c c}
\toprule
\multirow{2}{*}{System} & \multicolumn{3}{c}{Data configuration} & \multicolumn{3}{c}{Computational cost} \\
\cmidrule(lr){2-4} \cmidrule(lr){5-7}
 & Observation & Trajectory & Trajectory & Batch & Total & Training \\
 & dimension & number & length & size & epochs & time\\
\midrule
SOUP & 1 & $3\times 10^5$ &1000 & 1800 & 1200 & 5.8 h\\
OUP (fGn+L{\'e}vy) & 1 & $3\times 10^5$ & 1000 & 1800 & 2500 & 12.5 h\\
DVDP & 1 & $5\times 10^5$ & 500 & 1200 & 2000 & 9.7 h\\
CFHN ($\mathcal{O}_{x_1x_2}$) & 2 & $3\times 10^5$ & 1500 & 1200 & 3000 & 23.1 h\\
CFHN ($\mathcal{O}_{x_1}$) & 1 & $3\times 10^5$ & 1500 & 1200 & 3000 & 22.0 h\\
CFHN ($\mathcal{O}_{\bar{x}}$) & 1 & $3\times 10^5$ & 1500 & 1200 & 3000 & 23.0 h\\
Airfoil & 2 & $3\times 10^5$ & 1000 & 1800 & 5000 & 25.4 h\\
\bottomrule
\end{tabular}
\end{table}

\begin{figure}[!htb]
\center{\includegraphics[width=1\textwidth]
{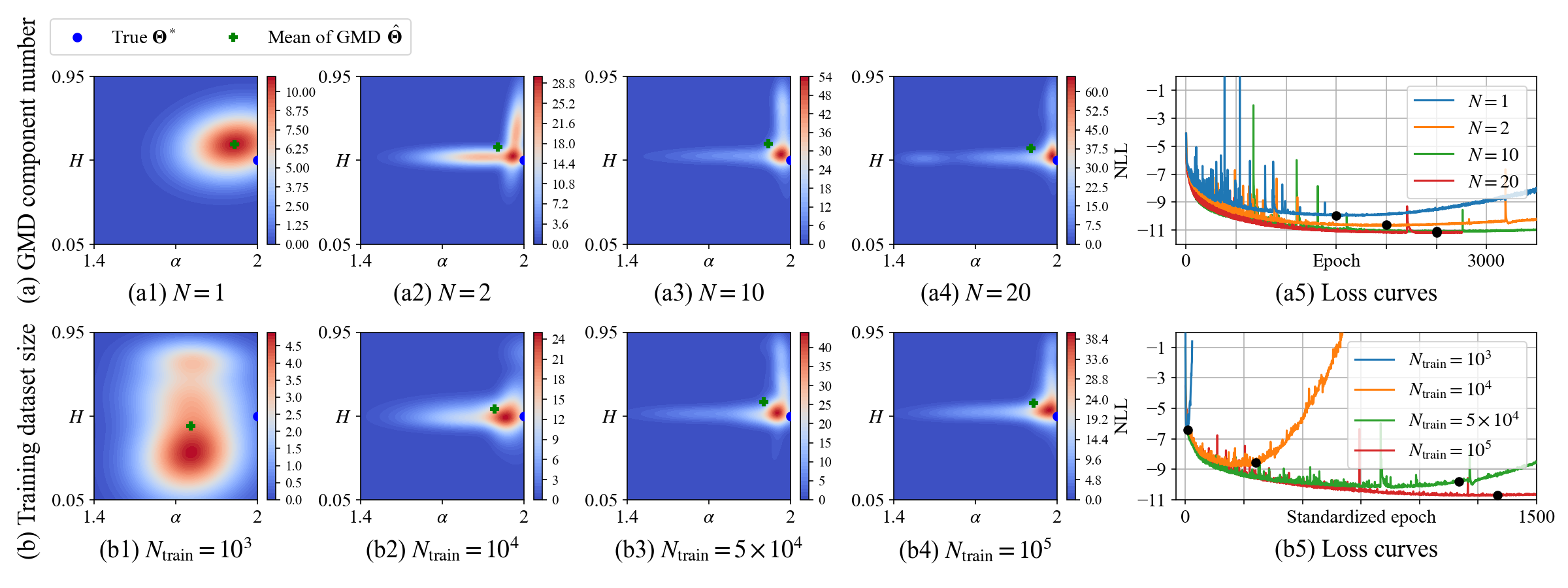}}
\caption{\label{fig:ablation}A comparison of PENN-GMDs with (a) $N=1$, $2$, $10$, and $20$ Gaussian components included in the GMD or (b) $N_\text{train}=10^3$, $10^4$, $5\times 10^4$, and $10^5$ trajectories included in the training dataset. The first four columns show the $(\alpha, H)$ likelihood distributions of a test trajectory with true parameters $\lg\tau=0.5$, $H=0.5,\alpha=2$, $D_{\mathrm{FGN}}=0.2$, and $D_\mathrm{L\acute{e}vy}=0.8$. The last column shows the loss curves of the 8 PENN-GMDs on a validation set with 20,000 trajectories. The black dots indicate the models for testing. In (b5), each standardized epoch processes $3\times 10^5$ trajectories.}
\end{figure}

The ablation studies of the encoder and decoder of PENN-GMD were extensively investigated in~\cite{Feng2023Deep, Feng2025Fusing}. Here, we investigate the number of Gaussian components $N$ and the size $N_\text{train}$ of the training dataset. By taking the fGn and L{\'e}vy-driven OUP system in Sec.~\ref{sec:oup} as an example, we have trained four PENN-GMDs with $N=1$, 2, 10, 20 Gaussian components and another four PENN-GMDs with $N_\text{train}=1000$, 10,000, 50,000, and 100,000 training trajectories. Tab.~\ref{tab:ablation_error} lists the quantitative metrics summarized on 20,000 test trajectories with uniformly sampled parameters. Figure~\ref{fig:ablation} shows the marginal likelihood distributions of $(\alpha, H)$ on a test trajectory and the loss curves.

When the component number $N$ increases from 1 to 20, Tab.~\ref{tab:ablation_error}(a) shows the mean absolute errors of the point estimates of the five parameters only decrease slightly and the SDs of the absolute errors are almost identical. Therefore, different Gaussian components lead to almost identical point estimates. This invariance is expected because the point estimate is the first moment (weighted mean) of the GMD, which approximates the posterior mean under the NLL loss. Increasing the number of components primarily enhances the capacity to resolve covariance structures and multi-modality, while the overall center of mass of the distribution remains largely unaffected and is already well captured even with a small number of components. In contrast, the last column of Tab.~\ref{tab:ablation_error}(a) shows that when the component number $N$ increases from 1 to 10, the mean NLL of the GMD estimates decreases significantly from $-5.370$ to $-6.495$, and the SD of the NLLs also decreases considerably from $3.039$ to $1.974$. This clear improvement shows that using 10 Gaussian components results in more accurate and precise GMDs that better capture the couplings and uncertainties of the trajectories. Further increasing the component number from 10 to 20, the mean and SD of the NLLs only decrease slightly, indicating that the accuracy is saturated, justifying our choice of $N=10$ in the baseline algorithm. 

\begin{table}[!htb]
\scriptsize
\centering
\caption{The mean/SD statistics of several PENN-GMDs with (a) different number of Gaussian components $N$ and (b) different training data sizes $N_\text{train}$ on 20,000 test trajectories of the fGn and L{\'e}vy-driven OUP system. The middle five columns show the absolute errors of the point estimates of the five parameters. The right column shows the NLL of the true parameters in the GMD estimate.}\label{tab:ablation_error}
\begin{tabular}{ccccccccc}
\toprule
Group & $N$ & $N_\text{train}$ & $|\lg\hat{\tau}-\lg\tau^*|$ & $|\hat{D}_\text{FGN}-D_{\mathrm{FGN}}^*|$ & $|\hat{H}-H^*|$ & $|\hat{D}_\mathrm{L\acute{e}vy}-D_\mathrm{L\acute{e}vy}^*|$ & $|\hat{\alpha}-\alpha^*|$ & $\ln q(\boldsymbol{\Theta}^*;\boldsymbol{\Phi})$\\
\midrule
\multirow{4}{*}{(a)}& 1 & \multirow{4}{*}{$3\times 10^5$} & 0.086/0.082 & 0.139/0.113 & 0.062/0.064 & 0.127/0.107 & 0.093/0.076 & -5.370/3.039\\
& 2  & & 0.086/0.081 & 0.138/0.114 & 0.060/0.064 & 0.126/0.107 & 0.094/0.074 & -6.043/2.625\\
& 10 & & 0.085/0.081 & 0.139/0.113 & 0.061/0.064 & 0.126/0.107 & 0.093/0.075 & -6.495/1.974\\
& 20 & & 0.085/0.081 & 0.138/0.113 & 0.060/0.064 & 0.126/0.107 & 0.092/0.075 & -6.558/1.928\\
\midrule
\multirow{4}{*}{(b)}& \multirow{4}{*}{10} & $10^3$ & 0.172/0.144 & 0.207/0.137 & 0.178/0.128 & 0.187/0.136 & 0.131/0.087 & -1.851/2.577\\
& & $10^4$ & 0.113/0.105 & 0.160/0.122 & 0.083/0.084 & 0.156/0.119 & 0.107/0.081 & -4.027/3.570\\
& & $5\times 10^4$ & 0.093/0.087 & 0.145/0.116 & 0.066/0.069 & 0.134/0.110 & 0.097/0.077 & -5.626/2.373\\
& & $10^5$ & 0.089/0.084 & 0.141/0.115 & 0.063/0.066 & 0.129/0.108 & 0.095/0.077 & -6.107/2.118\\
\bottomrule
\end{tabular}
\end{table}

Comparing the marginal distributions in Figs.~\ref{fig:ablation}(a1)-(a4), one can find that a single Gaussian distribution ($N=1$) in Fig.~\ref{fig:ablation}(a1) only forms an overly uncertain distribution. The two Gaussian components in Fig.~\ref{fig:ablation}(b1) roughly capture the main coupling patterns among the noise parameters $\alpha$ and $H$. However, the positive coupling when $\alpha$ is close to 2 is incorrect. The distributions in Figs.~\ref{fig:ablation}(a3) and (a4) that use $10$ and $20$ components are comparable and correctly capture all the coupling patterns. Despite this, we believe that in real applications, the best $N$ depends on the systems and may require an ablation study to decide, and 10 components is a good baseline setting.

When the size of the training dataset $N_\text{train}$ increases from $10^3$ to $10^5$, Tab.~\ref{tab:ablation_error}(b) shows that the point estimates of the five parameters are more accurate and precise. The mean NLL also decreases significantly, indicating the uncertainties are captured more accurately with a larger training dataset. Figs.~\ref{fig:ablation}(b1)-(b4) show that PENN-GMDs trained on $N_\text{train} \leq 5\times 10^4$ trajectories fail to precisely capture the T-shape likelihood distribution of the stability index and the Hurst exponent. More training data indeed improve the learning of the uncertainties and couplings among parameters.

The loss curves of the 8 PENN-GMDs listed in Tab.~\ref{tab:ablation_error} on a validation set with 20,000 trajectories are plotted in Figs.~\ref{fig:ablation}(a5) and (b5). All the loss curves decrease steadily first, indicating again that the training process of PENN-GMD is stable. When $N\leq 2$ or $N_\text{train}\leq 5\times 10^4$, the loss curves increase when the number of training epochs is large enough, implying that overfitting occurs. These results indicate that increasing the number of components improves the accuracy of uncertainty estimation, and that a sufficiently large training set is essential for achieving a good approximation of the population-level objective in Eq.~(\ref{eq:emp_and_kld}).

\begin{figure}[!htb]
\center{\includegraphics[width=1\textwidth]{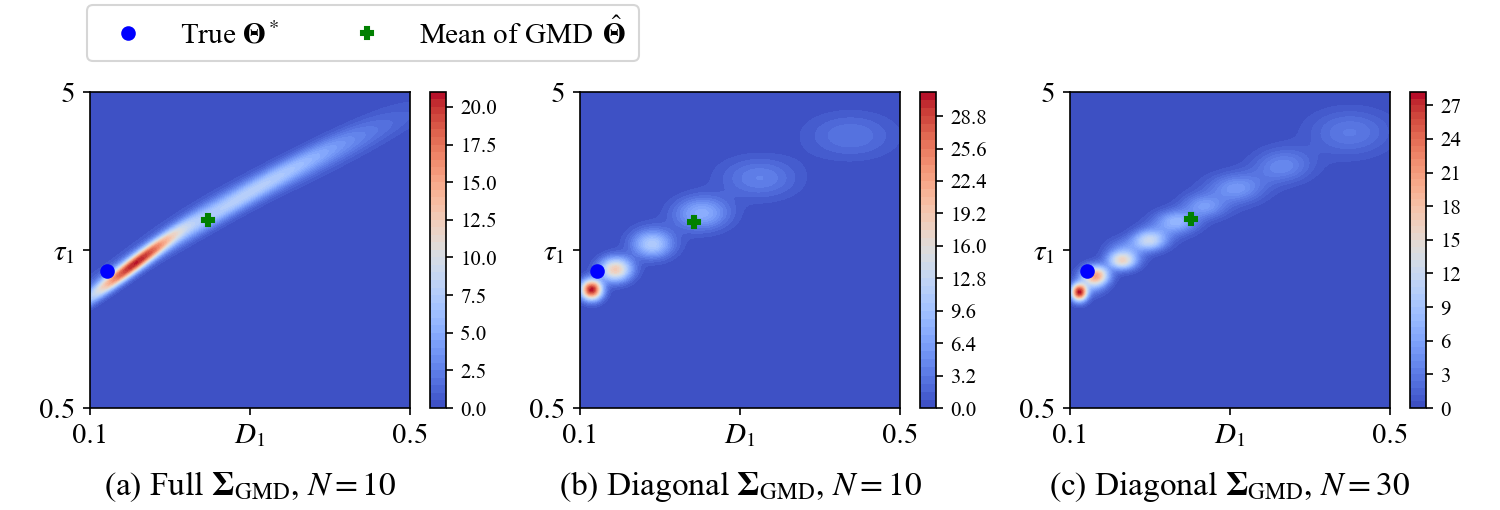}}
\caption{\label{fig:likelihood_dvdp_cov_type}Marginal likelihoods of the coupled parameters $D_1$ and $\tau_1$ for the DVDP trajectory in Fig.~\ref{fig:dvdp_trajectories}(b), estimated by PENN-GMDs with diagonal or full covariance Gaussian components.}
\end{figure}

To assess the practical advantage of full covariance over diagonal covariance in the GMD formulation, we train two additional PENN-GMDs on the DVDP system with $N=10$ and $N=30$ diagonal-covariance Gaussian components, keeping all other settings identical to Sec.~\ref{sec:dvdp}. Figure~\ref{fig:likelihood_dvdp_cov_type} compares the inferred marginal likelihoods of the strongly coupled parameters $D_1$ and $\tau_1$ for the trajectory shown in Fig.~\ref{fig:dvdp_trajectories}(b). With full covariance and $N=10$ (panel a), the likelihood is smooth and narrowly confined along the correlation ridge. With diagonal covariance, even $N=30$ components (panels b and c) produce only scattered axis-aligned masses, indicating that diagonal covariance cannot efficiently represent strongly coupled high-dimensional parameter spaces. This qualitative difference is reflected in the NLL. As reported in Tab.~\ref{tab:acc_dvdp_cov_type}, the full-covariance GMD with $N=10$ achieves a significantly lower NLL with the mean value $-9.357$ than its diagonal counterparts with $N=10$ ($-8.529$) and $N=30$ ($-8.833$), confirming that full covariance captures parameter uncertainties more effectively even with fewer components.

\begin{table}[htbp]
\footnotesize
\centering
\caption{Mean and SD of the NLL for the DVDP system with full-covariance and diagonal-covariance GMDs.}
\label{tab:acc_dvdp_cov_type}
\begin{tabular}{ccc}
\toprule
Covariance type & Number of components & $\ln q(\boldsymbol{\Theta}^*;\boldsymbol{\Phi})$ \\
\midrule
Full & 10 & $-9.357/3.721$ \\
Diagonal & 10 & $-8.529/3.646$ \\
Diagonal & 30 & $-8.833/3.590$ \\
\bottomrule
\end{tabular}
\end{table}

\section{Conclusion}
\label{sec:conclusion}
We have developed PENN-GMD, a neural framework that learns a closed-form Gaussian mixture approximation to the intractable likelihood of mixed-noise SDEs from single trajectories. Its key novelty lies in the end-to-end prediction of full-covariance GMDs with hard-encoded constraints, enabling direct extraction of parameter couplings and multi-modal uncertainties.

Through five progressively challenging numerical examples, namely a standard Ornstein-Uhlenbeck process with known likelihood, an OUP driven by mixed fractional Gaussian and L{\'e}vy noises, a Duffing-van der Pol oscillator with additive and multiplicative colored noises, a coupled FHN system under different observability scenarios, and a pitch-plunge aeroelastic airfoil with mixed-noise disturbances, we have demonstrated that the GMD output not only provides accurate point estimates but, more importantly, reveals the rich coupling structure and uncertainty landscape of the parameters. In cases of complete non-identifiability, such as the degeneracy of fGn and L{\'e}vy noise or the exchangeability of two neurons under mean-only observation, the PENN-GMD naturally splits into multiple modes that faithfully represent the manifold of plausible parameter combinations. In the airfoil case, even when the noise parameters are largely unidentifiable due to their weak and indirect influence, the GMD covariance matrix still captures the underlying physical compensation mechanisms among structural parameters, a diagnostic capability that is inaccessible to conventional point estimators or diagonal covariance models.

Despite its demonstrated effectiveness, the proposed framework has two limitations that warrant further investigation. First, the component number $N$ is fixed to 10. An adaptive selection strategy would be a valuable extension. Second, the full covariance matrices scale quadratically with the parameter dimension, increasing training cost for high-dimensional systems. These technical details, however, do not undermine the core validity of our approach, but rather point to practical directions for future engineering refinement.

\section*{Acknowledgements}

This study was partly supported by the NSF of China (Grant No. 52225211), and the National Natural Science Foundation of China (Grant Nos. 12202255, 12102341, and 12072264).





\normalem
\bibliographystyle{unsrt}
\bibliography{reference}

\end{document}